\documentclass[10pt,journal,transmag]{IEEEtran}
\ifCLASSOPTIONcompsoc
  \usepackage[nocompress]{cite}
\else
  \usepackage{cite}
\fi
\ifCLASSINFOpdf
\else
\fi
\IEEEoverridecommandlockouts
\usepackage[utf8]{inputenc}
\usepackage{mathptmx}  
\usepackage[hyphens]{url}  
\usepackage{hyperref}
\usepackage{graphicx} 
\usepackage{caption} 

\usepackage{cite}
\usepackage{amsmath,amssymb,amsfonts}
\usepackage{mathrsfs}
\usepackage{empheq}
\usepackage{textcomp}
\usepackage{subfig}
\usepackage{xcolor}
\usepackage{multirow}
\usepackage{booktabs}
\usepackage{array}
\usepackage{romannum}
\usepackage{lineno}

\newcommand{\ie}{i.e.}

\def\BibTeX{{\rm B\kern-.05em{\sc i\kern-.025em b}\kern-.08em
    T\kern-.1667em\lower.7ex\hbox{E}\kern-.125emX}}

\begin{document}


\title{Self-Growing Spatial Graph Network for \\
Context-Aware Pedestrian Trajectory Prediction
}

\hyphenation{op-tical net-works semi-conduc-tor}

%
%
%
%

\author{Sirin~Haddad,~\IEEEmembership{Member,~IEEE,}
        Siew-Kei~Lam,~\IEEEmembership{Senior Member,~IEEE}
        
\IEEEcompsocitemizethanks{\IEEEcompsocthanksitem S. Haddad was with the School
of Computer Science and Engineering, Nanyang Technological University (NTU), Singapore, 639798.\protect\\
E-mail: see siri0005@e.ntu.edu.sg
\IEEEcompsocthanksitem S.K. Lam is an Assistant Professor in NTU.}
\thanks{Manuscript received Dec 19, 2020.}}

\IEEEtitleabstractindextext{%

\begin{abstract}
Pedestrian trajectory prediction is an active research area with recent works undertaken to embed accurate models of pedestrians social interactions and their contextual compliance into dynamic spatial graphs.
However, existing works rely on spatial assumptions about the scene and dynamics, which entails a significant challenge to adapt the graph structure in unknown environments for an online system. 
In addition, there is a lack of assessment approach for the relational modeling impact on prediction performance.
To fill this gap, we propose \textit{Social Trajectory Recommender-Gated Graph Recurrent Neighborhood Network (STR-GGRNN)}, which uses data-driven adaptive online neighborhood recommendation based on the contextual scene features and pedestrian visual cues. The neighborhood recommendation is achieved by online Nonnegative Matrix Factorization (NMF) to construct the graph adjacency matrices for predicting the pedestrians' trajectories.
Experiments based on widely-used datasets show that our method outperforms the state-of-the-art. Our best performing model achieves 12 cm ADE and $\sim$15 cm FDE on ETH-UCY dataset. The proposed method takes only 0.49 seconds when sampling a total of 20K future trajectories per frame.
\end{abstract}

}
\maketitle

\IEEEdisplaynontitleabstractindextext

%
\IEEEpeerreviewmaketitle

\section{Introduction}
\label{sec:intro}
\footnote{This research was accomplished while being a PhD student at NTU.\\
Code and Data are available on Github: \href{https://github.com/serenetech90/AOL\_ovsc}{\textcolor{red}{here.}}\\
Additional Materials and Training files: \href{https://drive.google.com/file/d/1LDIUrK7vntxySmeSDXhXTEBIi4xjcD7P/view?usp=sharing}{\textcolor{red}{here.}}}
\IEEEPARstart{P}{edestrian} trajectory prediction plays an essential role in Advanced Driver-Assistance Systems (ADAS), autonomous driving, and robotic navigation to maintain pedestrians safety. Modern trajectory prediction approaches
\cite{van2018relational,bartoli2017context,fernando2018soft+} model social interactions locally, where the neighborhood boundaries are defined by a specific metric (i.e., short distance). An exception to this is Social Attention \cite{vemula2018social} and Social GAN \cite{gupta2018social}, which model all pedestrians interactions to create a global context. The local methods observe dynamics within the nearby pedestrian whereabouts and use a tabulated format to model the environment. 

Earlier research defined local pedestrian neighborhoods based on fixed spatial distances as in \cite{helbing1995social,alahi2016social,cheng2018pedestrian}. Forming neighborhoods neurally is already explained in \cite{yang2018my,battaglia2016interaction}. 
There are few works that developed the social modeling on spatio-temporal graphs considering the spatial relations between pedestrians  \cite{haddad2019situation,zhang2019sr, xue2018ss} using fixed parameters to determine the neighborhood boundaries. Such local capture considers related pedestrians, but with limited understanding of the social interaction temporal evolution.

Research in crowd behavioral modeling has reported a self-organizing cooperative tendency 
among individuals in the crowd \cite{helbing2005self,schiermeyer2016genetic,johora2018modeling,battaglia2018relational}. Starting with this intuition, we theorize a data-driven self-growing mechanism that learns the interaction between pedestrians. 
\begin{figure}
    \centering
    \includegraphics[width=0.9\linewidth, height=4cm]{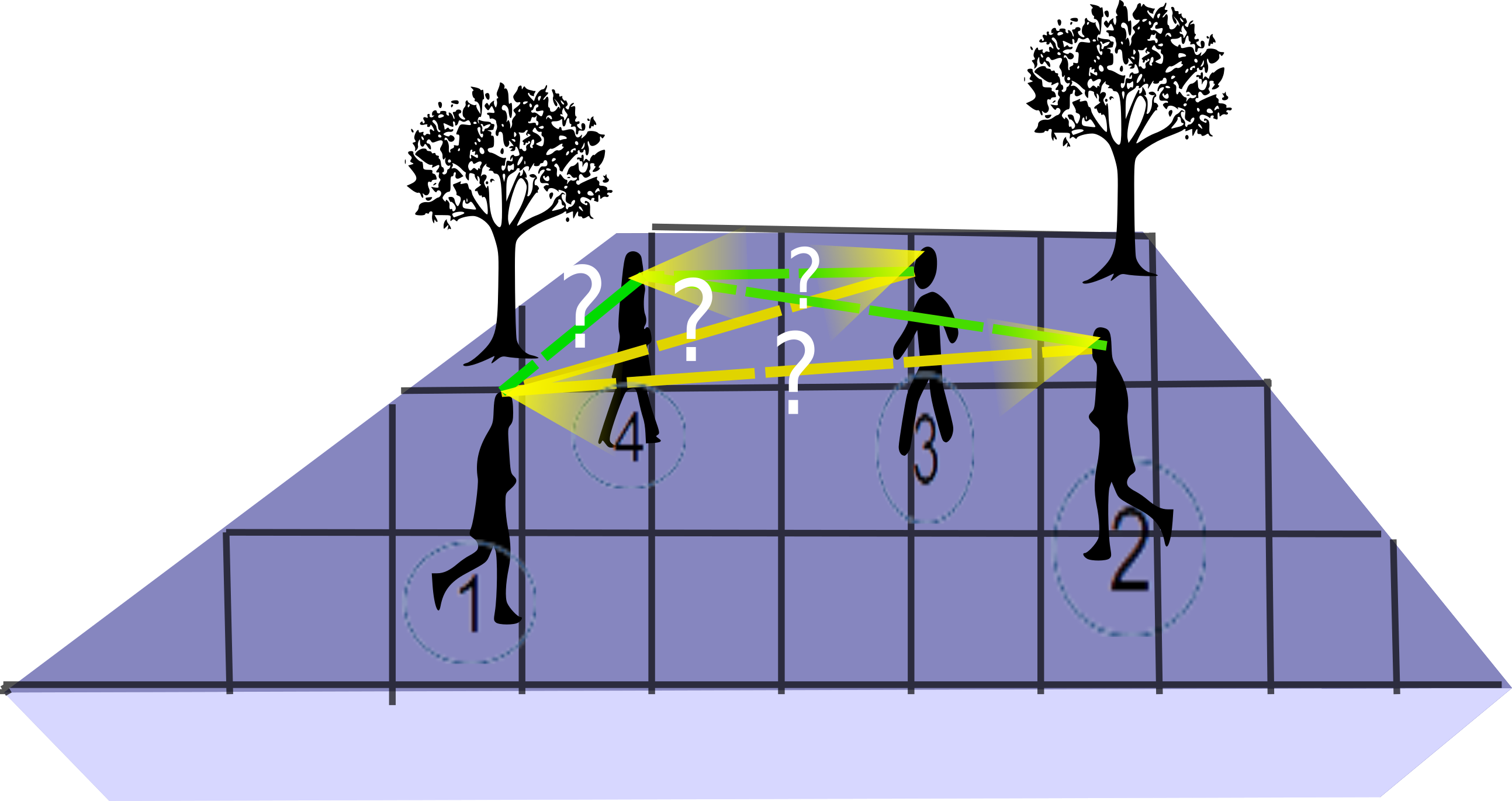}
    \caption{An illustration of the real-world scenario of pedestrians in an urban environment. Anticipating whether pedestrians are influencing each other is depicted by the $<$?$>$, and the dashed links connecting pedestrians indicate whether a potential relationship exists. Each pedestrian has a walking trajectory and looking span indicated by the light cones at the head.}
    \label{fig:intro_fig}
\end{figure}
Figure \ref{fig:intro_fig} visualizes the case where this mechanism is required for developing associations between pedestrians. By considering their visual span, one can be more certain about how pedestrians consider each others when moving. Accordingly, a learning mechanism can learn to estimate their impact and interactions. However, using a data-driven self-growing approach introduces additional challenges to model the social interaction on spatio-temporal graphs. There is a need to evaluate the generated neighborhoods plausibility as the graph grows neighborhoods dynamically. 
To overcome this problem, we employ pedestrians' visual cues as a means to capture their spatial awareness for constraining their neighbourhood.
We propose an approach that combines tabulated modeling with graph-structured modeling of interactions. Then it examines the impact of associating pedestrians on the effectiveness of social modeling, such that related pedestrians become pooled into adaptively-shaped neighborhoods. Moreover, we propose a self-learned relational inference that plays a significant role in growing the graph edges. At the same time, we dedicate a tabulated structuring for modeling the static context into local neighborhoods. This hybrid approach aims to improve the modeling of pedestrians social influence and their awareness of static surroundings, thereby reducing trajectory prediction errors.
Our framework employs GridLSTM \cite{kalchbrenner2015grid}, which is a more advanced LSTM cell that is capable of structuring its memory gate over multimodal streams of features simultaneously. GridLSTM passes memory states between data modalities horizontally and vertically, which enables it to learn faster and more effectively than LSTM-based models.
In summary, the contributions of our work are as follow: 
\begin{itemize}
\item We introduce STR-GGRNN, an edge-centric online framework for mapping crowd dynamics onto spatio-temporal graph networks and automatically infer the social interactions by completing the graph edges with minimal engineering effort.
\item Proposing a novel kernel to maps the STR-GGRNN features gradients and accordingly, generates the future trajectories as an effect of the interactions change relative to the static scene over time. 
\item To the best of our knowledge, our work is the first to integrate Nonnegative Matrix Factorization (NMF) as an efficient self-learned social neighborhood recommendation system. It neurally evaluates the importance of the edges with a compact version of nodes and edges features. The system generates variational neighborhood proposals and examines them against the prediction accuracy. It yields a locally optimal solution for modeling the social interactions in spatio-temporal graphs. 
\item The proposed model outperforms state-of-the-art methods in the ETH-UCY dataset using TrajNet++ challenge data \cite{sadeghiankosaraju2018trajnet}.
\end{itemize}

\section{Related Works}
\subsection{Pedestrian Trajectory Prediction}
Trajectory prediction approaches can be categorized into classical engineered models \cite{trautman2010unfreezing,robicquet2016forecasting, hasan2018seeing} and learning-based models. Encoder-Decoder architecture has achieved notable success in real crowd surveillance and driving datasets \cite{liang2020garden}. Examples include GAN-based network \cite{sadeghian2018sophie, gupta2018social,hasan2019forecasting, li2020social}, LSTM-based \cite{robicquet2016learning,fernando2018soft+, ma2019trafficpredict, zhang2019sr}, Gated Graph-structured networks \cite{jain2015car, jain2016structural,alahi2016social,vemula2018social,haddad2019situation,   ivanovic2019trajectron, kosaraju2019social, casas2019spatially, chandra2019forecasting,li2020evolve, haddad2020self,mohamed2020social}. In several occurrences, LSTM was combined with CNN in an end-to-end pipeline. This has shown to be effective in generating contextually compliant paths.  
The LSTM-based and other methods used prior knowledge and empirical practices of modeling pedestrian trajectory and hence, they are tailored to specific scenes and datasets that contains fewer pedestrians and homogeneous scenarios captured by ETH-UCY dataset\cite{lerner2007crowds}.


\subsection{Evolutionary Graph-based Systems Modeling}
Fundamental research proposed biologically-inspired techniques to model complex dynamic systems on a graph as a set of nodes connected by a set of edges. Graph-based modeling had a plethora of successful applications, and with numerous advancements this modeling evolved to adopt biological evolution processes into dynamically adaptive graphs that encode features and make decisions about structural growth. Neural Gas \cite{fritzke1995growing} and self-Organizing Maps \cite{kohonen1990self} introduced the fuzzy self-learned growth into graph theory.

Recently, in the field of crowd motion modeling, there are several attempts to model crowd holistic motion as a simulation of a physically-inspired system \cite{xie2017learning,karasev2016intent, battaglia2016interaction, battaglia2018relational,yau2020graphsim}. However, in such proposals the underlying graph/temporal function was resembling a pre-engineered natural model and the growth decision was determined or predicted a-priori. 

Self-organization and hierarchical networks in artificial intelligence root back to Self-Organizing Maps \cite{kohonen1990self} and Graph Neural Networks (GNN) \cite{ranganathan2003self,  helbing2005self,scarselli2008graph, perez2014self, wu2020comprehensive}. These promoted the growth of network hierarchy from its own learning and inference process. The uncertainty of the association between two nodes creates a necessity for an adaptively-formed edge set. Such a proposal entails more flexible modeling of complex variable-sized environment and dynamic behaviors. This idea was theorized over a more general context in \cite{smith2009growing} as an evolutionary fuzzy graph clustering by assigning degrees to indicate relationships between similar nodes. 

This work adopts the self-learning graph concept and introduces it into graph-based pedestrians modeling to propose more powerful graphs that can adaptively change, and create its structure throughout a physics kernel that computes the higher order of interactions features to stem their evolution. The kernel accurately and efficiently model context-aware interactive pedestrians. Considering that graph growing is a fuzzy process and graph completion is a NP-hard problem, the objective here is introduce a variational graph growth to choose the local optimum solution with minimum modeling of mechanical functions for pedestrians motions and intents.

\subsection{Efficient Graph-based Modeling}
With predefined-neighborhood settings, \cite{alahi2016social, cheng2018pedestrian, xue2018ss, li2019grip} predictors surmised that pedestrian neighborhoods cast on a fixed grid. Moreover, they used hardcoded proxemic distance for outlining neighborhood boundaries. The family of global graph Structural-RNN \cite{jain2016structural}, Social Attention \cite{vemula2018social}, SpAGNN \cite{casas2019spatially}, and Fuzzy Graph Attention \cite{kamra2020multi} grow fully-connected graphs with degree-based relationships to cluster pedestrians. They assume neighborhoods to be the global context of the scene and start with non-fuzzy edge set but end with relational estimation that assign different degrees to the edges, creating neurally-evaluated fuzzy edge sets to recognize related pedestrians. 
Similarly, interaction graphs \cite{battaglia2018relational,li2020evolve, kosaraju2019social, li2020social, ma2019trafficpredict} make neural relational inference (NRI \cite{kipf2018nri}) given the heterogeneous agent dynamics. 
In light of these methods, we allow a greater degree of freedom in the pedestrians social-spatial modeling by instantiating self-learned proxemic-free criteria for saving the additional parameter tuning and engineering effort.
Contrary to interaction graphs, our approach stores compact features in the graph. It performs neural restructuring of neighborhoods without attributing the edges. It just encodes past and present motion sequences for each pedestrian and replaces the initial edge-set with sparse adjacency matrix. That is, the objective is to re-generate more efficient graphs by means of minimizing edges without the need for an exhaustive search through all the possible solutions.

\subsection{Graph-Structured Data Streams}
Recently, Adaptive Online Learning has emerged as a new trend in pedestrian trajectory prediction. \cite{huynh2020aol} reports prediction results on portions of the ETH-UCY datasets. We aim to advance the application of online learning over multiple modalities of data. Unlike offline training, online learning can be more sensitive to the catastrophic interference problem and requires a large memory space. Data streaming application requires immediate processing and mapping in lightweight structures. Graph kernels are useful techniques for critical online prediction systems that model and generate predictions with continuous evaluation of pedestrian dynamics and relationships to the environment objects \cite{yau2020graphsim}.

This work shows how mapping the graph to kernel can be a valid technique for achieving online continuous learning of the scene and pedestrians motion features in order to track their and stem future pedestrian motion relevant to the scene structure.

\subsection{Link Prediction through Graph Recommender Systems}

Link Prediction techniques estimate the relations that tie graph nodes to each other \cite{feng2020atbrg,li2013recommendation,mutinda2019time,shi2020heterogeneous,fan2019graph}. Link prediction is eminent in recommender systems, social networks, and biologic-ally-inspired growth networks \cite{chen2005link, tran2017community, liben2007link, Lei2012novel}.
The link prediction quality can vary based on the goodness of nodes and edges features. In the worst-case scenario, the Breadth-First Search (BFS), takes up to quadratic time $\mathcal{O}(n^2)$ time to traverse an offline graph with $n$ nodes and complete the missing edges. 
This time complexity is even more challenging in online graph streams to estimate relational heuristic over an incomplete graph. Theoretically, there is a lack of a globally optimal solution to the right neighborhood discretization, and there is no prior-stated universal edge set. 
As explained previously, the current online application \cite{hua2020online,huynh2020aol,bera2014adapt} is limited to few pedestrians due to the large memory consumption. 

Our work is the first to combine online paradigm with social interaction inference on incomplete spatio-temporal graphs. We coin this approach as Self-Growing Online Graph. The pedestrian social inference problem is tackled as an online matrix completion problem that employs variational unsupervised technique called Nonnegative Matrix Factorization (NMF). This factorization carries out a compact representation of nodes relationships through the online graph. 
We postulate that a Self-Growing model can flexibly encode any mixture of social and contextual features besides facilitating a scalable network growth that we will show in our experimental results. 


\section{Proposed Approach}
\subsection{Problem Formulation}

Given a set of pedestrians $\mathcal{N}$, we represent their trajectories over time $T^i_{t}$ using a spatio-temporal graph $G_t$ at each time step $t$, containing \ensuremath{\mathcal{N}} nodes. $T^i_{t}$ comprises the 2D positional trajectories $X^i_{t} = (x^i_t, y^i_t)$ and 2D head pose sequence $V^i_{t} = (v^i_{x_t}, v^i_{y_t})$. The i-th pedestrian is assigned a node $n^i_t$ to store the ground-truth trajectory $x^i$, the 2D head pose $v^i$, and temporal edge $e_t$ that links the node $n^i_{t-1}$ to $n^i_t$. We propose $P$ adjacency matrices $A$ for all pedestrians at each time-step and predict i-th future trajectory $\widetilde{X} = {\widetilde{x}^{iA^1}_{t+1}, \widetilde{x}^{iA^2}_{t+2}, ..., \widetilde{x}^{iA^P}_{t+l}} $ per each matrix $A^p$. The best adjacency matrix that produces the minimum prediction errors is then selected. 
$A^P = {A^1, A^2, ..., A^p}$ and $\widetilde{x}^{iA^p}_{t+l}$ is the future trajectory of length {t+l} steps in the future, for i-th pedestrian in the p-th Adjacency matrix proposal, given the past observed trajectory between $x^i_{t}$ to $x^i_{t-obs}$, where $obs$ is the observation length, including the current step. 

\subsection{STR-GGRNN Framework}
\label{sec:approach}
\textit{STR-GGRNN} is a framework for processing pedestrians interactions and trajectories as graph data streams. It includes crowd data mapping into spatio-temporal graph $G_t$, and then models their spatial relationships by completing the graph spatial edges $e_t$ at time-step $t$. 

The framework takes pedestrians positional trajectories $X$ and encodes them using a nested linear transformation function $\phi$, such that $W_X$ is [8 x 10] and $W_{\hat{X}}$ is [10 x 8]. Eventually the trajectories are encoded as [10 x 10] as a lightweight version:

\begin{equation}
    \hat{X} = \phi (X)
\end{equation}
\begin{equation}
    \phi (X) = W_{\hat{X}} * (W_X * X)
\end{equation}

It also encodes the 2D head pose annotations, called Vislets, $V$ using a single transformation function $\phi$. The lightweight encoding size is [2 x 10], where $W_V$ is [2 x 8]: 
\begin{equation}
    \hat{V} = \phi (V)
\end{equation}
\begin{equation}
    \phi (V) = W_V * V 
\end{equation}

The head pose is annotated in the ETH-UCY datasets in roll-yaw format. This is to represent the head rotation at the back-forth axis and along the vertical axis. The 2D poses were annotated in the radian system which describes the angular change in head direction, such that it would be more relevant to modeling pedestrians attention and easier for training the network compared to the angular degree system.

Our framework then concatenates the embedded cues $\hat{V}$ and $\hat{X}$ at the vertical axis and feed T into GGRNN:
\begin{equation}
    T = (\hat{X}, \hat{V})
\end{equation}{}

The proposed spatial-temporal graph network adopts an edge-centric task, as the graph analytics focus on predicting associations (edges) between pedestrians. It is a central model since a single GridLSTM cell is used as the encoder of the social pedestrians interactions.
It takes graph $G_t$ of $\mathcal{N}$ nodes and $\mathcal{E}$ edges, and maps it into a kernel of fixed dimension, $K$, such that the kernel generates predicted trajectories $\widetilde{X}$ and adjacency states between the pedestrian nodes. The adjacency models the social interactions. 
\begin{equation}\label{eqn:kernel}
    \widetilde{X} , J_\theta = K(G_t{(\mathcal{N},\mathcal{E})}, H_t^*)
\end{equation}{}

The Social Trajectory Recommender (STR) component is a recommender system that performs matrix factorization to complete the Adjacency matrix.The framework encodes the contextual influence of static features onto pedestrian trajectory by segmenting the scene into a grid of local regions which comprises the Gated Graph Recurrent Neighborhoods Networks (GGRNN). These regions hold the potential contextual interactions between pedestrians and the scene structure. They are assigned continuously over time, creating a sequence of graphs encoded by gated Recurrent Neural Networks. Overall, the framework ends up combining the contextual model features from GGRNN and the pedestrians social interaction features from STR. 


We will first discuss in depth the GGRNN and explain its formalization and task. Later, we will present the formulation of the Social Trajectory Recommender component (STR) and illustrate how the different components are integrated into the GGRNN pipeline.

\subsection{Gated Graph Recurrent Neighborhood Network (GGRNN)}
\begin{figure*}
    \centering
    \includegraphics[width=\linewidth, height=6cm]{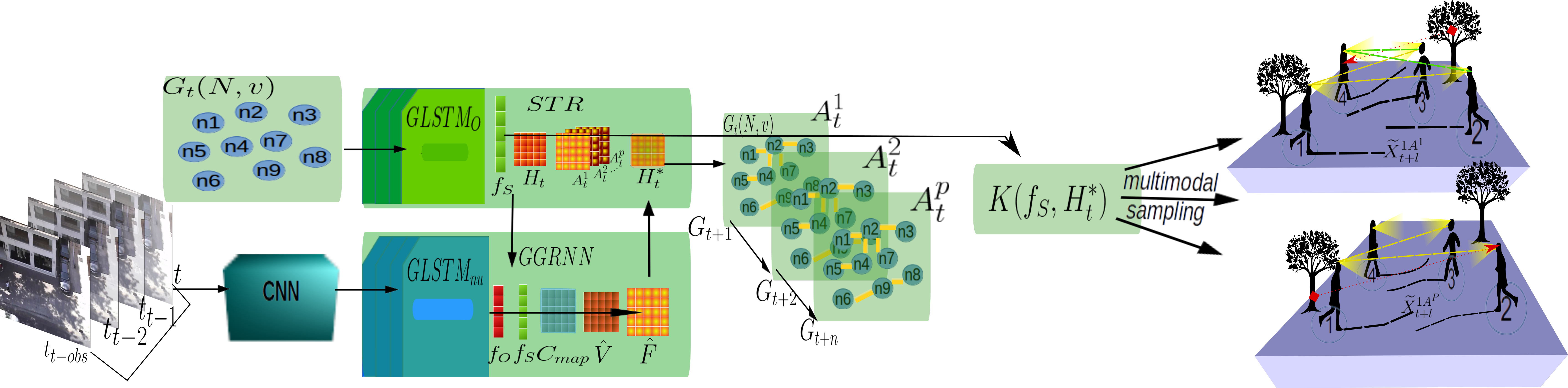}
    \caption{The full online pipeline of STR-GGRNN. The STR network encodes Vislets $\hat{V}$ and positional trajectories $\hat{X}$ for each pedestrian trajectory. Then maps them into social grid $f_S$ using $GLSTM_{nu}$. The GGRNN network discretizes static context using $GLSTM_O$ into 'Visuospatial' neighborhoods and stores pedestrian contextual awareness in $f_O$. At the consequent step, STR takes $f_O$ and $f_S$ and maps them into the weighted adjacency matrix using the NMF technique. It will generate the edge set $\mathcal{E}$ as means of completing graph $G_t$ spatial edge-set while the temporal edge-set \{$G_{t+1}$, $G_{t+2}$, $G_{t+3}$\} connect the spatial instances over time. The concept of Gated Graph is depicted to the right-side such that set of temporal graphs has a gated edge completion process according to GridLSTM gating mechanism.}
    \label{fig:STR}
\end{figure*}{}
The Gated Graph Recurrent Neighborhood Network takes the static scene and encodes it as a fixed local neighborhoods grid. It models the contextual interactions between pedestrians and their context surrounding. This contextual model contributes to generating future trajectories that account for pedestrians "context-awareness" to make plausible predictions.

Firstly, it takes the transformed Vislets $\hat{V}$ and trajectories $\hat{X}$ from the transformation functions. 
It assumes that motion happens over an initially uniformly-divided square grid containing $v$ static neighborhoods. The STR$^*$ variants run single GridLSTM cell $GLSTM_{\nu}$ on each graph $G_t$ which is encoding features $T$ and initial hidden states $h_0$, to generate relative social features $f_S$:
\begin{equation}
    \label{eqn:ngh_glstm}
    f_S, h_S = GLSTM_{nu}(T, h_0)
\end{equation}{}

In Figure \ref{fig:STR}, $CNN$ is a 2D convolutional layer used for encoding the human-space interaction that takes a static scene image as input. A grid mask $M$ is applied to the convoluted features. The filter depicts static space neighborhoods that are discretized as a square grid and is uniformly-initialized.
\begin{equation}
  C_{map}  = CNN(f_S, h_S) * M
\end{equation}{}

In our work we encode head pose $V$ with static features $C_{map}$ to formulate the "Visuospatial" neighborhood representation using single GridLSTM cell $GLSTM_O$:
\begin{equation} \label{eqn:visuospatial}
     f_\mathcal{O}, h_\mathcal{O} = GLSTM_O(C_{map}, \hat{V}, h_O)
\end{equation}{}

The usage of GridLSTM as a multimodal feature encoder is clarified in (\ref{eqn:visuospatial}). It combines visual awareness state $V$, the scene spatial feature map $C$, and the normally-initialized hidden states $h_O$, for stemming pedestrians attention to the physical context. It stores pedestrians contextual interaction in $f_\mathcal{O}$. 

Eventually, GGRNN produces a final neighborhood representation, $F$, combines the outputs of GLSTM$_\nu$ with 

GLSTM$_O$, which are static features grid with social features grid, static scene map and Vislets again:
\begin{equation}
    F = f_S * f_{\mathcal{O}}^\ensuremath{\prime}
\end{equation}


\subsection{Social Trajectory Recommender (STR)}
\label{sec_str}
In this section, we formalize the proposed model for recommending social trajectories given a band of estimated neighborhood proposals. We coin variational modeling as Social Trajectory Recommender. The proposals suggest several unique edge sets to connect pedestrian nodes. Then it selects the best neighborhood proposal, which generates minimum prediction errors, among other proposals.

Figure \ref{fig:STR} shows the detail steps for the recommendation mechanism to achieve deep relational inference between pedestrians.
The proposed mechanism has a customized design relevant to the features set included in our model versions. 
The kernel $K$ generates the social interaction states $H_{t+1}$ and future positional predictions $\widetilde{X}$. Based on STR-GGRNN and ST-V model variants, the kernel uses $f_\mathcal{O}$ and $f_{S}$ to calculate each pedestrian adjacency state. As such, Eq. (\ref{eqn:kernel}) becomes:
\begin{equation}
   \label{mc_mcr_K}
    \widetilde{X}, H_{t+1} = K(f_S, V, H_t^*)
\end{equation}{}

Here, $K$ calculates the soft attention over social and contextual interaction features at one hand, and calculate their gradient in relevance to the static map at the other hand:
\begin{gather}
    \label{mc_mcr_K_layers}
    \hat{F} = \phi(F) = C_{map} * ( (W_v * [f_S, \hat{V}] + b_v) * (W_r * F) )\\
    J_{\theta} = (\frac{d (f_S * W_v || f_\mathcal{O})}{df_\mathcal{O}})\\
    J_{\theta} = ReLU (J_{\theta})\\
     \widetilde{X} = MLP(J_{\theta}) = W_o * (W_c * J_{\theta})
\end{gather}
$\hat{F}$ holds a final neighborhood representation. It combines the outputs of GLSTM$_{nu}$ with GLSTM$_O$, which are static features grid with social features grid, static scene map and Vislets again. This ensures a rich multi-modal feature representation and strengthens the intertwining between contextual and social interactions. 
But according to our basic model $LSTM_O$ and ST-GGRNN, the kernel $K$ in  Eq. (\ref{eqn:kernel}) has a limited formulation that only generates future trajectories without the social recommendation or contextual features, hence this kernel version takes parameters as follows: 
\begin{equation}
    \label{glstm_K}
    \widetilde{X}, H_{t+1} = K(f_S, H_t^*)
\end{equation}
In the variants that exclude social inference, $K$ calculates the soft attention over social and contextual features at one hand in the same way as previous variants,however, it extrapolates future trajectories from convoluting static map with social neighborhoods to create a relevance effect between pedestrians social interactions and their interactions with the scene features:
\begin{gather}
    \hat{F} = \phi(F) = C_{map} * ( (W_v * [f_S, \hat{V}] + b_v) * (W_r * F) )\\
    J_{\theta} = (W_v * F + b_v) * C_{map}\\
    J_{\theta} = ReLU (J_{\theta})\\	
    \widetilde{X} = MLP(J_{\theta}) = W_o * (W_c * J_{\theta})
\end{gather}

After generating future trajectories, $\hat{F}$ now is used to compute scaled soft-attention weights. STR model will tune the influence of each region to update contextual features and evaluates the social relationships strength. It deploys the scaled self-attention mechanism \cite{velickovic2017graph} (as illustrated in Eq. (\ref{eqn:soft_attn}). Attention coefficient $a$ considers the human-human interaction features and the human-space interaction features:
\begin{equation} \label{eqn:soft_attn}
    a =  \frac{Softmax (\exp{(\hat{F})})}{\Sigma \exp{(\hat{F})}}
\end{equation}{}

Given the $i$th pedestrian trajectory features, neighborhood $nu^i_t$ of i-th pedestrian has 1 or more j pedestrians considered as neighbors to i:

\begin{equation}
\label{eq:nu_def}
    nu^i_t =  \{(i, j)\}^+; \quad |nu^i_t| <= |\mathcal{N}|
\end{equation}{}

Moreover, neighborhood boundaries at pedestrian $i$ are defined by the set of edge pairs that connect pedestrian nodes to other nodes in the graph.
Online graph completion is an NP-hard problem, so we rely on a stochastic adjacency matrix generation mechanism that generates one adjacency proposal at a time-step. During back-propagation, it evaluates the estimated adjacency proposal with all the possible permutations for that crowd. It optimizes the objective function responsible for selecting neighbors based on minimizing the L2-Norms resulted from the best adjacency matrix.
All the proposals yielded from Eq. \ref{eq:nu_def} for adjacency matrix belongs to the solution space $\Omega$:
\begin{equation}
    \label{eq:omega_space}
    \Omega_{nu_t} = \{nu^i_1, nu^i_2, ..., nu^i_{nxn}\}
\end{equation}
Selecting the best adjacency will lead to a local optimum solution for the neighborhood construction, as the best optimum is not known, and a brute-force approach towards testing all the possible permutation is of exponential cost to the deployment. So our method considers $\Omega$ a partially-observed variable and resorts to the non-exhaustive graph search mentality.

The static features $f_\mathcal{O}$ are weighted by soft-attention coefficient $a$ using the soft-attention mechanism in \cite{bahdanau2014neural}: 
\begin{equation}
    f_{\mathcal{O}_{t+1}} = a * f_{\mathcal{O}_{t+1}}
\end{equation}{}

The attention weightage manifests pedestrian relationships have various degrees of strength.
Inspired by the neural factorization technique \cite{webb2019factorised} to factorize spatio-temporal graph edges, we establish a deep mechanism for neighborhoods that is aware of the static and the social constraints.
Again, the best setting for neighborhoods is not fully-observable. We estimate pedestrian relationships weight based on relative visual attention and relative spatial motion features to define their neighborhoods. 
We introduce the Nonnegative Matrix Factorization (NMF) into our pipeline for a proper approximation of social relations from the neural attention features and static map. The reconstruction yields a more compact and efficient adjacency representation to be carried to the next time-step, such that $A_0$ is the initial $A$ matrix (all ones) of size [10 x 10] for 10 pedestrians in the scene, $W_a$ is assigned attention $a$, $H_t$ is assigned the weighted hidden states:
\begin{equation}
   W_a, H_t = \textit{NMF}(a,C_{map} C_{map}^\top )
\end{equation}

\begin{equation}
    A^P = W_a / max(W_a)
\end{equation}

$W_a$ is a square matrix of the frame max-size (\ie  max number of pedestrians). We select $W_a$ as it approximates the motion features in its weighted version and normalizes it to 1. $W_a$ maps directly to adjacency states of range [0,1], as the edge weights are nonnegative values.

The NMF reconstructs a compact version of the input features to approximate the social interaction patterns between pedestrians. Using the downscaled features enhances the scalability within an online pipeline. 

Finally, within STR$^*$ recommender models we apply Softmax function to node hidden states. This will transform nodes states into continuous space under the interval [0,1]:
\begin{equation} \label{eqn:rcmndr_fnri}
     A_t;\quad min(||\quad X_i - \widetilde{X}^{iA^P} \quad||_2)
\end{equation}{}

We display in (\ref{eqn:adj_all_models}) the formal definitions of the proposed mechanisms for autonomous relational modeling, including ours (\ie STR$^*$): 
\begin{equation}
\label{eqn:adj_all_models}
     A_t = 
     \begin{cases}
         1/(|| x_i^t - x_j^t ||_2); \quad M_{SGTV} \\
          Softmax (W*H_t); \quad MCR^* \\
         min(||\quad X_i - \widetilde{X}^{iA^P} \quad||_2) ; \quad STR^* \\
    \end{cases}
\end{equation}

The Self-Growing model (SGTV) \cite{haddad2020self}, which was the first proposed method for growing the edge set $SGTV$, relied on hard-attention weights to express the importance of relationships. The weights were calculated using pedestrian-wise Euclidean distances. However, their relational inference outcomes are unexpected, and the neighborhood modeling would change for each run. The ultimate degree of freedom found in such data-driven modeling makes it difficult to evaluate their modeling effectiveness.

Pedestrian states are updated with the selected best adjacency state to reflect upon the STR adjacency outcomes, $H^*_t$ gets passed to the following time-step:
\begin{equation}
    H^*_{t} = A_t*H_t
\end{equation}

\section{Experiments}

\subsection{Training Setup}

The learning objective is formulated as minimization of Euclidean errors between the p-th predicted trajectory $\widetilde{X}$ and the ground-truth trajectory $X$, given the locally-optimal adjacency proposal (edge set) $A^p_t$, such that there are P proposals:
\begin{equation} \label{eqn_costfn}
     \mathcal{L} = \underset{A^p_t}{argmin} ||\quad \widetilde{X_t} - X_t^p \quad||_2
\end{equation}

We empirically set the number of samples for variational trajectory and adjacency samples (P) to 10/20/100/200. We did not observe a change in the prediction accuracy. As an acceleration of our code, we deploy the variational parts over multi-threads to simultaneously sample future trajectories. Eventually, the code selects the minimum global Euclidean error based on the least erroneous predicted trajectories. Accordingly, the proposal with minimum Euclidean errors is the sub-optimal Adjacency matrix $A$, which sets the neighborhood boundaries for each pedestrian at time-step $t$ and produces the least prediction errors. 

The prediction errors are reported in meters over interpolated real-world coordinates in ETH-UCY dataset. We run experiments over two settings for observation/prediction lengths. In the first part, we observe 8 frames (3.2 seconds) and predict 12 steps (4.8) seconds. In the second part, we observed 4 frames (1.6 seconds) and predicted the next 8 frames (3.2 seconds).
We implement all model variants using TensorFlow (1.13) \cite{abadi2016tensorflow}. The experiments are deployed in a leave-one-out on desktop Intel® Xeon(R) CPU of frequency 3.50-GHz using Ubuntu 16.04 at a learning rate of 5e-3, decay rate of 0.95 and a dropout of 0.80. The maximum size variable (maxSize) of pedestrians is also set to 20 per frame. We repeated the testing experiments 10 times over each set to verify the errors stability. $GLSTM_o$ and $GLSTM_{nu}$ (outputSize) is 8 and embedding size is 128. Size of $h_S$ and $h_O$ is [20 x 128]. CNN encodes the static scene at compressed size [8x8] for compactness.

\subsection{Evaluation Metrics and Baselines}
Similar to \cite{hasan2018mx,alahi2016social}, we use the following Euclidean average errors:
\begin{itemize}
    \item Average Displacement Error (ADE):
          measures prediction errors along the time-steps between the predicted trajectory and the ground-truth trajectory as follows:
    \begin{equation}
          \frac{\sum_{i = 1}^N \sum_{j = 1}^l ||(\widetilde{X_i^j} - X_i^j)||_2 } {N * |T|} ,
    \end{equation}{}
    \item Final Displacement Error (FDE): measures prediction errors at the final time-step between the predicted trajectory and the ground-truth trajectory as follows:
    \begin{equation}
            \frac{\sum_{i = 1}^N ||(\widetilde{X_i} - X_i)||_2 } {N} \quad
    \end{equation}{}
\end{itemize}

\paragraph{\textit{Deterministic Models}}
\begin{enumerate}
    \item Trajectron++ \cite{salzmann2020trajectron++}: An edge-centric spatio-temporal LS-TM graph. We refer to their deterministic version. The model encodes contextual map and social interactions for heterogeneous dynamic agents of variably-sized environments. Therefore, its comparison with ST-GGRNN variant is considered valid. 
    \item \textit{S-LSTM} \cite{alahi2016social}: Dedicates LSTM for every pedestrian, and pool them given an empirical setting for the neighborhood.
    \item $LSTM_O$: Online GridLSTM model that only takes the positional trajectories $X$ with a fixed temporal window of 8 time-steps.
    \item GGRNN-V: This variant withdraws the social recommender section and includes only the static context neighborhoods. It highlights scenarios wherein pedestrians are not necessarily socialising (\ie posing direct impact on each other) or there is a decreased crowding. At its best, this variant models trajectories as an effect of reaching the target place and considering pedestrians visual awareness of the static surroundings.
    \item ST: This is the fully-connected graph model. It does not incorporate Vislets, static context neighborhood components or social recommendation
    \item ST-GGRNN: This follows the same pipeline of ST variant, with incorporation of the static context neighborhoods component (GGRNN).
    \item ST-V: This is the fully-connected graph model with Vislet being incorporated for each node. However, it does not incorporate either of the static context neighborhood components and social recommendation.
\end{enumerate}

\paragraph{\textit{Stochastic Models}}
\begin{enumerate}
    \item \textit{S-GAN} \cite{gupta2018social}: Social GAN, a variational Encoder-Decoder architecture for generating future predictions using GAN network.
    
    \item MX-LSTM \cite{hasan2018mx}: deployed 2D head poses in the model to encode the Visual Field of Attention (VFOA) in a cone-shaped looking span, such that the latter determines pedestrians visual attention scope and can substitute the looking angle feature. 

    \item Trajectron \cite{ivanovic2019trajectron}: A spatio-temporal graph for social modeling of variably-sized environments. The model predicts a band of trajectories and selects the best future trajectories the same as in the S-GAN model.
    
    \item \textit{Social-STGCNN} \cite{mohamed2020social}: A Graph Convolutional Network (GCN) that estimates relationships using a pedestrian clustering technique to assign weights given the euclidean proximity between pedestrians pairwise.
    
    
    \item STR-V: a model variation that applies social neighborhood selection knowing pedestrians location and their visual angle, $VS$ (\ie Vislets). This is to assess the plausibility of neighborhoods generated by the recommender technique.
    \item STR-GGRNN-V: By incorporating the Vislets, the model enriches the feature representation and forms a prior knowledge for the recommender technique so to assess the neighborhood selection with respect to the static context that draw its influence on pedestrians trajectory and social interaction.
\end{enumerate}
\begin{table*}
    \centering
    \caption{Analysis of the reported ADE/FDE errors across state-of-the-art models on ETH \& UCY video datasets. $\downarrow$ denotes decrease \% and $\uparrow$ denotes the increase \% in our models prediction ADE/FDE over Trajectron++. Respectively, STR-GGRNN-V achieves($\downarrow$84\%)/($\downarrow$85\%) while ST-GGRNN achieves($\downarrow$51.1\%)/($\uparrow$24\%). ST-GGRNN is the fully-connected graph, while STR-GGRNN-V is the recommender-based graph.}
    \resizebox{\linewidth}{!}{
    \begin{tabular}{c|ccccccccc}
    \hline
    &  \multicolumn{9}{c}{Stochastic models} \\ \hline
    Dataset & S-GAN & MX-LSTM & Trajectron & Trajectron++ & Social-STGCNN  & ST-GGRNN & STR-GGRNN & ST-GGRNN-V & STR-GGRNN-V \\\hline
    ETH-Univ& 0.81/1.5 & -- &  0.59/1.14 & 0.39/\textbf{0.83} & 0.64/1.11 & \textbf{0.29}/1.192 & 0.35/0.80 & -- & --\\
    Hotel& 0.72/1.61 & -- & 0.35/0.66 & \textbf{0.12/0.21} & 0.49/0.85 & 0.32/1.150 & 0.39/0.88 & -- & --\\
    Zara1& 0.34/0.69 & 0.59/1.31 & 0.43/0.83 & 0.15/0.33 & 0.34/0.53 & 0.31/1.154 & 0.47/0.95 & \textbf{0.10}/0.84 & 0.12/\textbf{0.14} \\
    Zara2& 0.42/0.84 & 0.35/0.79 & 0.43/0.85 & 0.11/0.25 & 0.30/0.48 & 0.30/1.160 & 0.40/0.91 & \textbf{0.10}/0.84 & 0.12/\textbf{0.143} \\
    Zara3& 0.45/\textbf{1.12} & --  & -- &  -- & -- & \textbf{0.31}/1.175 & 0.41/0.90 & -- & --  \\
    UCY-Univ& 0.66/1.44 & 0.49/1.12 & 0.54/1.13 & 0.20/0.44 & 0.44 / 0.79 & 0.27/1.630 & 0.40/0.92 & \textbf{0.10}/0.86 & 0.13/\textbf{0.15} \\ \hline
    AVG & 0.47/1.02 & 0.48/1.07 & 0.47/0.92 & 0.18/0.40 &  0.44/0.75 & 0.23/1.165 & 0.40/0.89 & \textbf{0.10}/0.85 & 0.123/\textbf{0.144} \\
    \hline
    \end{tabular} 
    }
    \begin{center}
    \begin{tabular}{c|cccc}
    \hline
    \multicolumn{5}{c}{Deterministic models} \\ \hline
    Dataset & LSTM$_O$ & Trajectron++ & S-LSTM & GGRNN-V \\\hline
    ETH-Univ & 1.76/3.87 & 0.71/1.68 & 0.50/\textbf{1.19} & \textbf{0.46}/1.55 \\
    Hotel & 0.81/1.78 & \textbf{0.22/0.46} & 0.24/0.56 & 0.45/1.56 \\
    Zara1 & 1.20/2.86 & 0.33/0.77 & \textbf{0.29/0.72} & 0.45/1.54 \\
    Zara2 & 2.25/5.14 & \textbf{0.23/0.59} & 0.27/0.67 & 0.47/1.51 \\
    Zara3 & 0.96/2.38 & -- & 0.84/1.80 & \textbf{0.48/1.24}\\
    UCY-Univ & \textbf{0.22/0.53} & 0.41/1.07 & 1.30/2.38 & 0.47/1.56\\
    \hline
    AVG & 1.06/2.44 & \textbf{0.38/0.92} & 0.66/1.36 & 0.47/1.50 \\
    \hline
    \end{tabular}
    \end{center}
    
    \label{tab:baseline_ours_test}
   
\end{table*}

\subsection{Quantitative Analysis}
Table \ref{tab:baseline_ours_test} lists existing state-of-the-art models ADE/FDE errors. We observed a general tendency across all models (Deterministic and Stochastic) to have the lowest ZARA subsets' errors and approximate errors over Hotel and ETH subsets. This tendency indicates that these models customize performance during the offline training to specific dynamic crowds patterns. 
GGRNN-V was used where Vislets $V$ annotation was available, \ie UCY and Zara. It performs similarly to GGRNN , with slightly higher ADE by 14\%. 
ST-GGRNN shows improvements over the former variants as it generates fully-connected spatial graphs. 

Fully-connected graphs are not be suitable for less interactive crowds, yet ST-GGRNN ADE outperforms Trajectron under ETH and Hotel. In contrast, it becomes comparable with the spatio-temporal baselines for the rest of the datasets. By observation, ETH and Hotel show crowds that conduct motion according to individual targets without persistent social interactions. Instead, such crowds are influenced by the static context. Noticeably, under Zara sets and UCY, the social and contextual factors impact pedestrians, so STR-GGRNN has a comparable performance with MX-LSTM given that the social relational inference does not take Vislets in this variant. That comparison leads us to verify the NMF-based relational inference, which designated a true capability of modeling interactions between pedestrians. 
Similar to ST-GGRNN, STR-GGRNN generates dense spatial graph with the visual cue and slightly improve ADE, however, the ADE/FDE discrepancy remains considerably large. 
STR-GGRNN-V achieves better improvements than STR-GGRNN, as the former reduces the metric discrepancy by lowering FDE significantly. The selective social inference procedure has better impact on the overall performance when there are additional cues that model pedestrian intention (\ie Vislets \& static map), and thereby STR-GGRNN-V delivers the best predictions over other spatio-temporal state-of-the-art.

Tables \ref{tab:sdd_desire_ours_new} and \ref{tab:sdd_ours_meters} compare STR-GGRNN best and overall performance errors along with the most recent state-of-the-art. Table \ref{tab:sdd_desire_ours_new} lists the methods in chronological order and illustrate pixels resolution. The most recent baselines such as Social-WaGDAT and Evolve-Graph reported their results on 1/5th resolution order, while STR-GGRNN was tested on SDD data with full pixel resolution and yielded comparable accuracy.
Additionally, to compare with STGAT and DAG-NET, STR-GGRNN is trained on SDD meter annotations. Under the pixels annotation STR-GGRNN shows fair improvements and under the meters annotation, it shows an outperforming improvements.
\begin{table}
	\caption{Comparing best results achieved in recent graph-based models with the proposed models at \textbf{full resolution}. The observation/prediction length in $M_{SGTV}$ is 1.67/3.2 seconds, while in the rest of methods it is reported at 3.2/4.8 seconds.}
	\begin{center}
		\begin{tabular}{c|c|c}
			\hline
			Model & Best & AVG \\ \hline 
			S-LSTM \cite{alahi2016social} & 9.85 & 33.2/56.4 \\ 
			Desire-SI-IT4 \cite{lee2017desire}& -- & 35.4/57.6 \\ 
			S-Attn \cite{vemula2018social} & -- & 33.3/55.9 \\ 
			CAR-NET \cite{sadeghian2018car} & -- & 25.72/51.80 \\
			GRE-MC-10 \cite{choi2019looking} & 22.35 & 27.0/43.9 \\ 
			$M_{SGTV}$ \cite{haddad2020self} & 9.81 & 24.46/59.53 \\
			NRI\cite{kipf2018nri} & -- & 25.6/43.7\\ 
			Trajectron++ \cite{salzmann2020trajectron++} & -- & 24.3/40.1 \\
			Social-WaGDAT \cite{li2020social} & -- & 22.52/38.27 \\
			Evolve-Graph \cite{li2020evolve} & -- & \textbf{15.3}/27.9 \\
			STR-GGRNN & \textbf{9.67} & 17.23/\textbf{24.24} \\ 
			ConvLSTM\cite{ridel2020scene} & -- & 14.35/26.85 \\\hline
		\end{tabular} 
	\end{center}{}
	\label{tab:sdd_desire_ours_new} 
\end{table}

\begin{table}[]
\caption{Comparing ADE/FDE results achieved in STR-GGRNN with the recent spatio-temporal graphs, the models are trained on Stanford Drone Dataset meters annotation \cite{sadeghiankosaraju2018trajnet}. }
	\centering
	\begin{tabular}{c|c|c}
		\hline
		Model & Best & AVG \\ \hline 
		STGAT\cite{huang2019stgat} & -- & 0.58/1.11\\ 
		Social-Ways\cite{amirian2019social} & -- & 0.62/1.16 \\
		DAG-NET\cite{monti2020dag} & -- & 0.53/1.04 \\ 
		STR-GGRNN & 0.04 & \textbf{0.05}/\textbf{0.07} \\ \hline
	\end{tabular}
	
	\label{tab:sdd_ours_meters}
\end{table}

\subsection{Qualitative Analysis}
\begin{figure}[h!]
\begin{center}
    \subfloat[]{\label{fig:all_pred_band}
    \includegraphics[width=4cm, height=4cm]{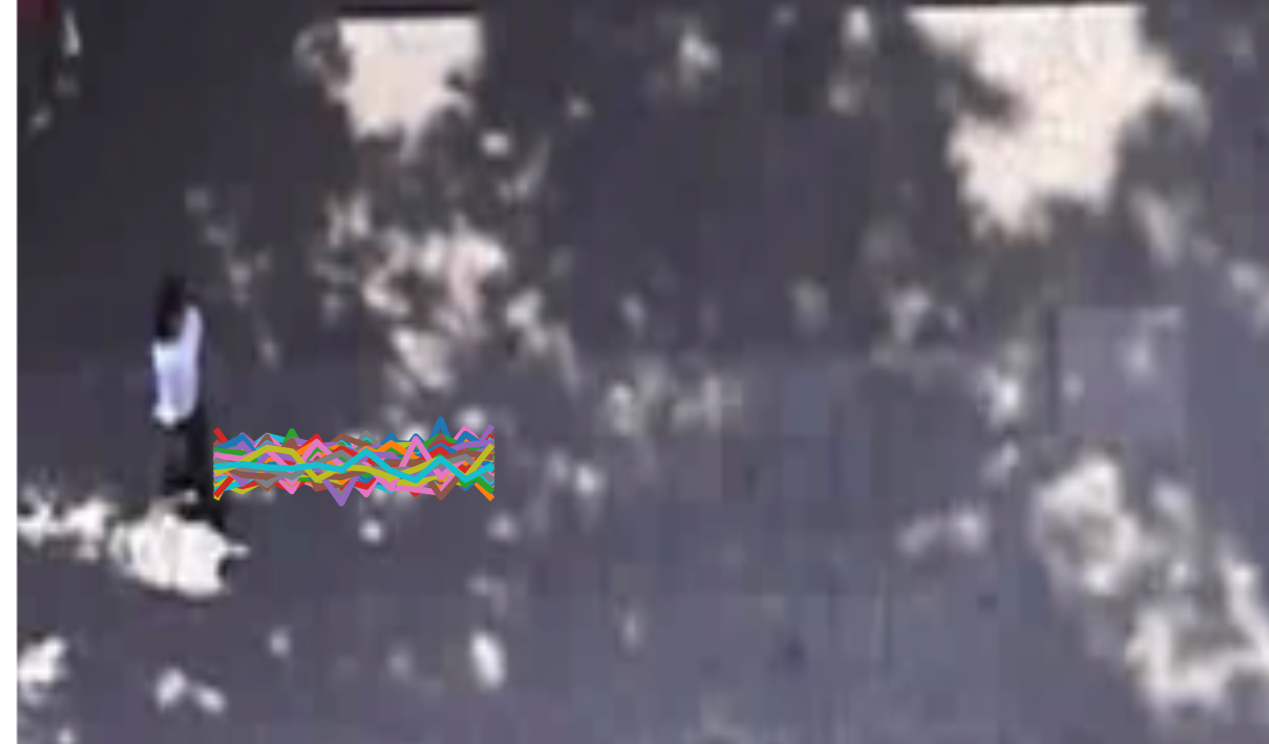}}
    \subfloat[]{\label{fig:frame_8177}
    \includegraphics[width=4cm, height=4cm]{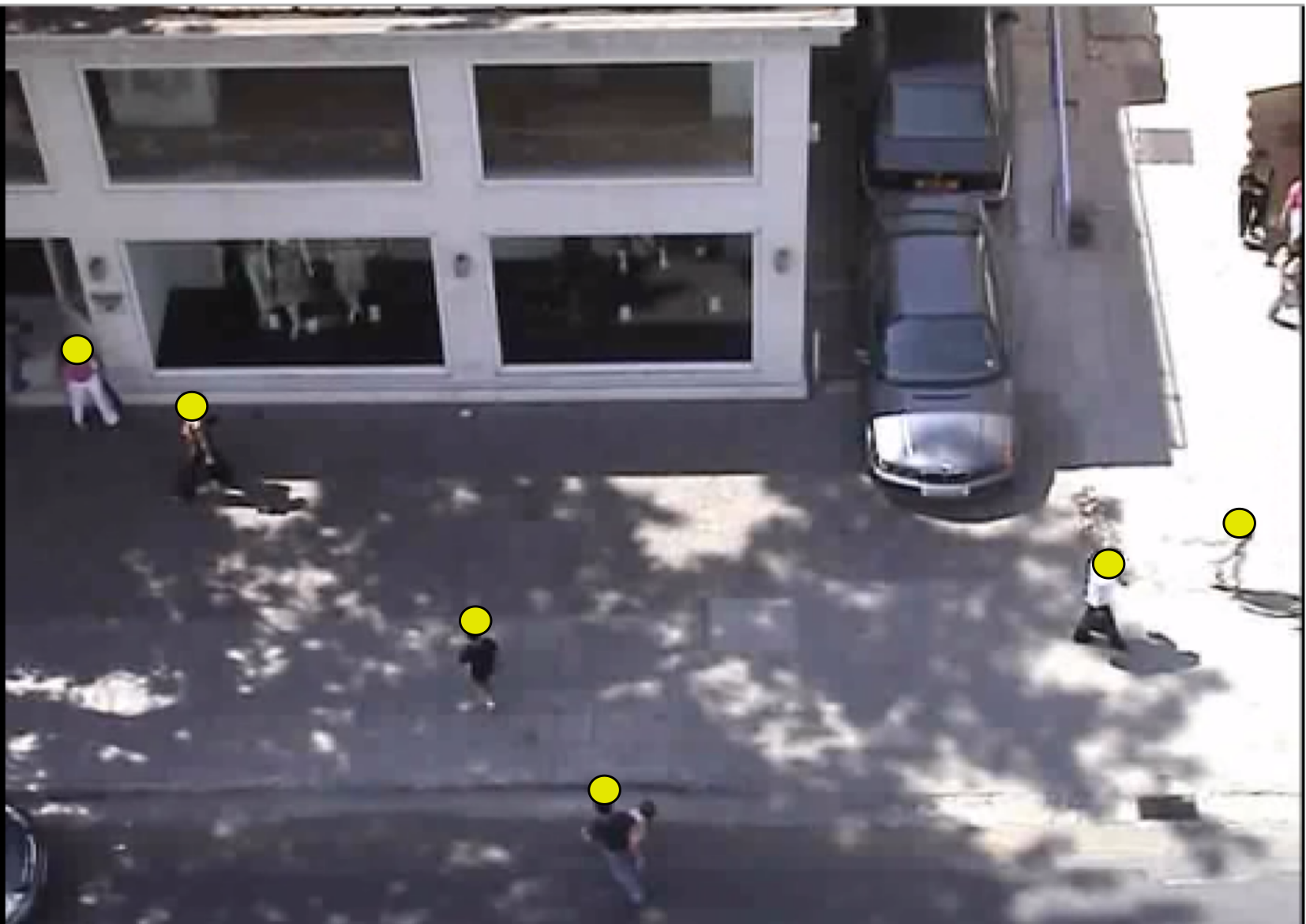}}
  
   \subfloat[]{ \label{fig:bst_adj_proposals}
    \includegraphics[width=4cm, height=4cm]{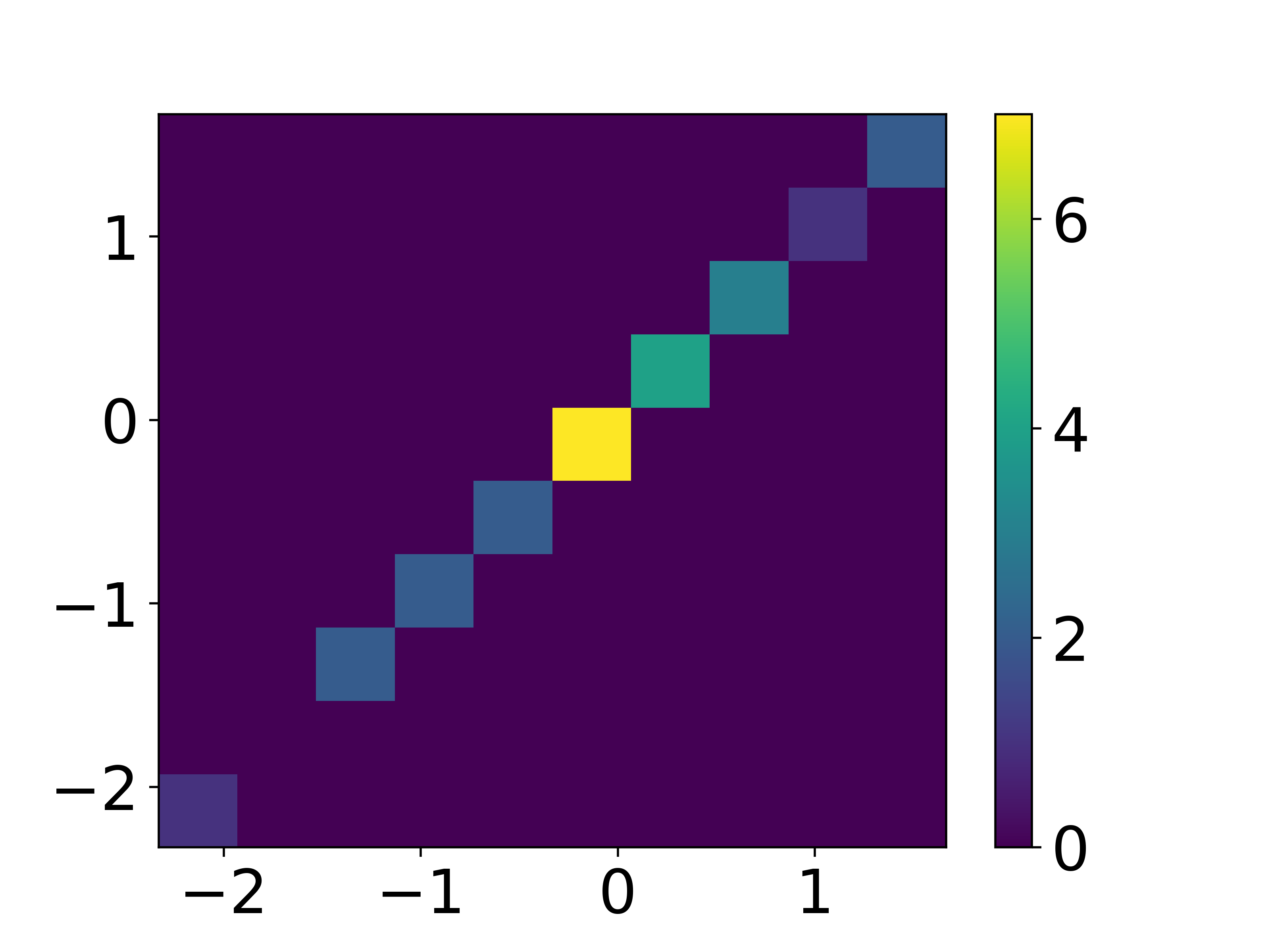}}
    \subfloat[]{\includegraphics[width=4cm, height=4cm]{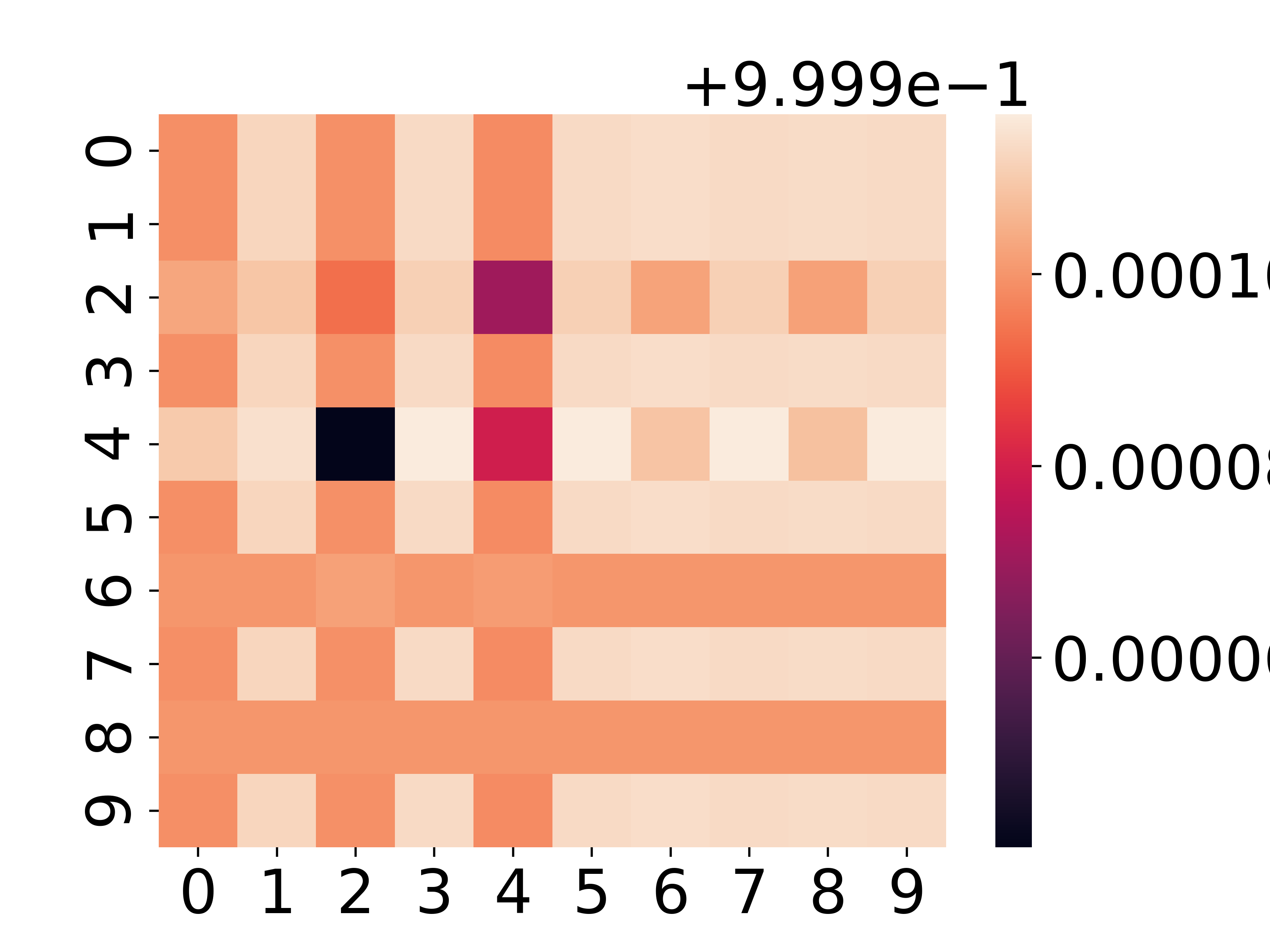}}
    
    \subfloat[]{\includegraphics[width=4cm, height=4cm]{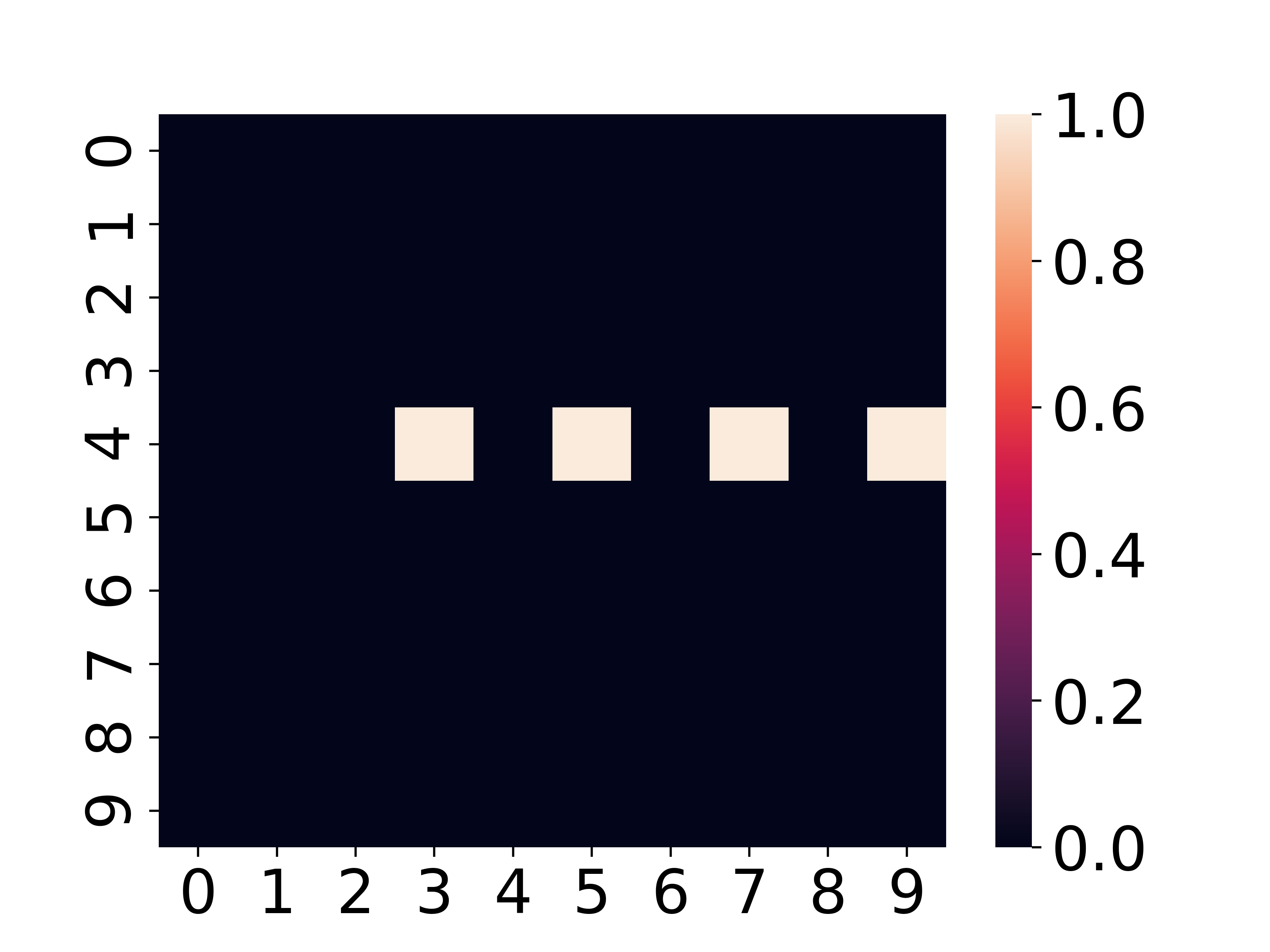}}
    \subfloat[]{\includegraphics[width=4cm, height=4cm]{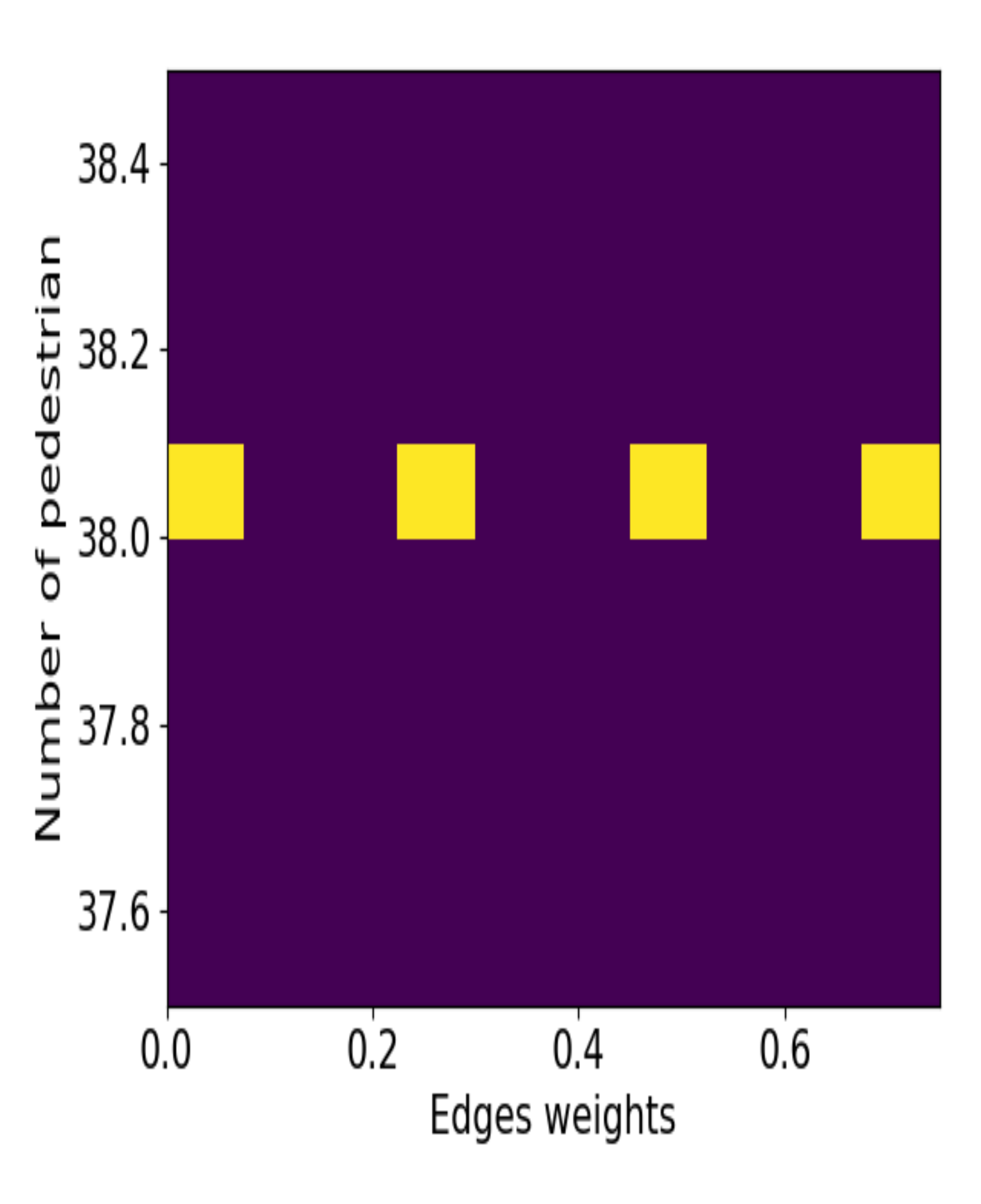}}
   \caption{\textbf{(a)} Prediction Bands: 100 sampled trajectories per pedestrian in frame (7961). \textbf{(b)}. Frame 8177 of Zara1 scene. \textbf{(c)} Adjacency proposal 2D histogram for frame in (a). \textbf{(d)} Adjacency proposal heatmap for frame in (b). \textbf{(e)} Maximum values in the adjacency matrix heatmap of (d), out of 6 pedestrians in (b), 4 pedestrians were taken as highly influential to each other. Those were in the proximity to Zara shop entrance. \textbf{(f)} Variations in weight estimation density for 38 pedestrians over 4 frames. The variations indicated that adjacency weights increased over time due to the increase in relational estimation confidence as the STR consequently modeled and mapped the social interactions to the adjacency matrix.}
\end{center}
\end{figure}
We visualize the same experiment several times to see the extent of variational bands. For clarity, Figure \ref{fig:all_pred_band} displays one band of 10 variational predicted trajectories for pedestrian (ID:128), in Zara1, frame (7961). The variational bands are compact and closer to the ground-truth trajectory than the variational prediction of S-GAN, which are more divergent.
Figure \ref{fig:bst_adj_proposals} visualizes the 2D histograms of the best-proposed adjacency under the same frame, containing only a single pedestrian. The selected proposal is shown in the figure. However, there are 10 2D histogram plots, each corresponds to [10 x 10] adjacency matrix $A^p_t$. As discussed before, the edges are assigned weights between [0,1] inclusively. The lighter shades represent higher weights, which means a higher degree of association strength. The yellowish squares at the line indicate a high density of connections. 

\section{Time and Space Complexity}

\begin{figure}
    \centering
    \includegraphics[width=\linewidth,height=6cm]{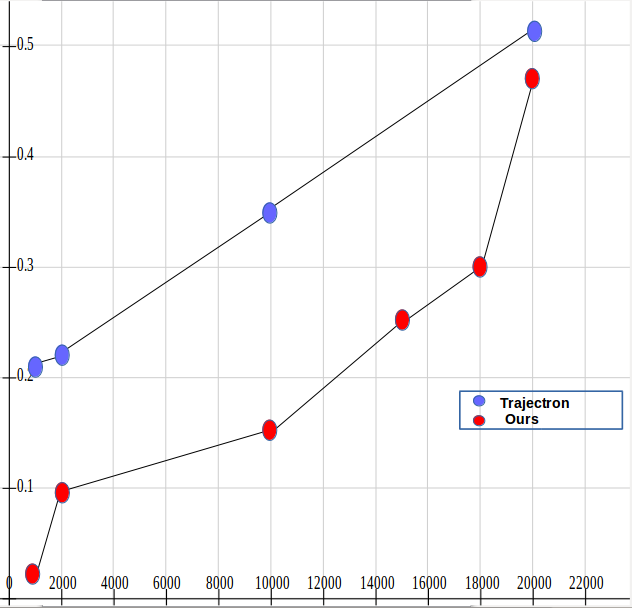}
     \caption{Sampling time curves of Trajectron \cite{ivanovic2019trajectron} vs. Our Kernel. The graph plot visualizes an approximate time complexity functions. The X-axis is the number of sampled trajectories and Y-axis is the sampling time in seconds.}
    \label{fig:timecomplex_curve}
\end{figure}

We observed a log-polynomial growth rate in both time and memory online training curves. This was the general pattern under both models. Figure \ref{fig:timecomplex_curve} visualizes the pattern in sampling time over different rates ranging from 1000 to 20K trajectories per single frame of 20 pedestrians. While Trajectron starts high, it yields a steadily linear running time. STR-GGRNN sustains logarithmic curve for sampling sizes up to 10K samples. Thereafter, the sampling time increases in a step-wise polynomial pattern such that it is asymptotic to Trajectron curve. 

The internal stages of training, such as generating adjacency proposals, takes $\sim$100 ms for STR (including the kernel $K$ sampling) to make 10 recommendations for a neighborhood of 10 pedestrians. It takes $\sim$360ms to choose the best proposal.
\begin{table}
\caption{Mean sampling time in our model Vs. Baselines best performing model over different sampling counts per frame, given that future trajectories are sampled at 12 steps ahead on Intel® Xeon(R) CPU E5-1650 v2 @ 3.50GHz with 16GB memory. Baseline models: Social GAN \cite{gupta2018social}, Social LSTM \cite{alahi2016social} and Trajectron \cite{ivanovic2019trajectron} were re-evaluated on our CPU to make fair comparisons.}
    \centering
    \resizebox{.95\columnwidth}!{
    \begin{tabular}{l|llll}
    \hline
    No. Sample & 1K & 2K & 10K & 20K \\ \hline
    S-GAN & 7.10s & 10.70s & 16.69s & 33.87s\\
    Trajectron & 0.22s & 0.23s & 0.36s & 0.52s \\
    S-LSTM & 8.07s & 15.67s & 78.37s & 155.67s\\
    S-Attn & 17.30s & 34.04s & 173.89s & 342.55s \\
    STR-GGRNN & \textbf{0.03s (86x)} & \textbf{0.10s (56x)} &  \textbf{0.15s (58x)} & \textbf{0.49s (5.7x)} \\\hline
    \end{tabular}}
    
    \label{tab:sampling time}
\end{table}
Table \ref{tab:sampling time} lists the sampling time required at the sampling stage.
At 20K trajectories per frame, the kernel sampling time in addition to recommending the best adjacency, become comparable to Trajectron, which means that our method can scale up by 100x more trajectories per entire frame. During initialization, our algorithm takes $\sim$3.54 seconds to start with the 1st batch. 

\begin{table}
\centering
\caption{Time and Memory space consumption summary for 8/12 steps (\ie observation/prediction lengths) and at sampling rate of 10K trajectories per frame on CPU.}
\resizebox{\columnwidth}!{
\begin{tabular}{c|ccc}
		\hline
		Model & Memory (MB) & Train Time (sec.) & Val. Time \\ \hline
		LSTM$_O$ & 422 & 2.30$\pm$[0.16, 1.00] & $\sim$2.26 \\
		STR-GGRNN & 500 & 2.90$\pm$1.00 & $\sim$2.30 \\\hline
\end{tabular}
}
\label{tab:spacetime_summary}
\end{table}
Table \ref{tab:spacetime_summary} summarizes the average statistics of time \& space occupied for a single run of the algorithm. 
The total time required to complete the social recommendation and evaluating the trajectories band for all pedestrians is $\sim$0.37 seconds with an interval of $\pm$ 0.10 seconds difference for several runs and 0.30-second increase from one batch to another while training. 

\section{Ablation Study}

\begin{table}[h!]
    \centering
    \caption{Ablation Study. Performance of various model components for 20 pedestrians in the scene.}
    \begin{tabular}{c|c|c|c|c}
    \hline
        Dataset & ST & STR & ST-V & STR-V\\
        \hline
        ETH & 0.38/1.26 & 0.40/1.23 & -- & -- \\
        Hotel & 0.38/1.29 & 0.38/1.22 & -- & -- \\
        Zara1 & 0.37/1.15 & 0.35/1.23 & 0.86/2.02 & 0.88/2.02 \\
        Zara2 & 0.36/1.24 & 0.37/1.29 & 0.89/2.02 & 0.88/2.02 \\
        Zara3 & 0.38/1.23 & 0.37/1.29 & -- & --\\
        UNIV & 0.41/1.29 & 0.36/1.23 & 0.86/2.02 & 0.88/2.02\\
        \hline
        AVG & 0.38/1.24 & 0.37/1.25 & 0.87/2.02 & 0.88/ 2.02 \\
    \hline
    \end{tabular}
    \label{tab:ablation_table}
\end{table}

To verify the components impact on the overall performance, ablative experiments were performed to observe different features impact on the social modeling effectiveness. Noticeably, running STR component performs better given only the positional trajectories as in STR and ST than it does when combined with head poses. STR-V model performed much worse than model STR-GGRNN which combines static context. This illustrates that the head poses alone were not sufficient feature composition for generating effective social recommendation, unless it is combined with the static context map. This means that STR was more capable of modeling individuals social interactions given their visual awareness to the static context. The head pose was more beneficial when being used as an indicator of the static environment impact than when used as pedestrians influence indicator. In fact, visual span represents pedestrian observability of other pedestrians, yet, it does not form a robust basis for estimating pedestrians attention to each other unless this observability is combined with motion features. 

\section{Graph Efficiency Analysis}
A thorough statistical analysis has been conducted to visualize STR algorithm behavior compared to fully-connected graph (fullGT) and Trajectron approach.

\begin{figure}
    \centering
    \subfloat[25th/75th Whiskers Box Plot SGTV vs. STR]{\label{fig:whisker_plt}
    \includegraphics[width=\linewidth,, height=5cm]{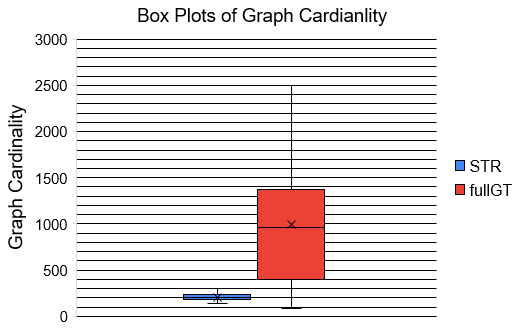}
     }
     
    \subfloat[25th/75th Whiskers Box Plot Trajectron vs. STR]{\label{fig:whisker_plt_1}
    \includegraphics[width=1.02\linewidth, height=5cm]{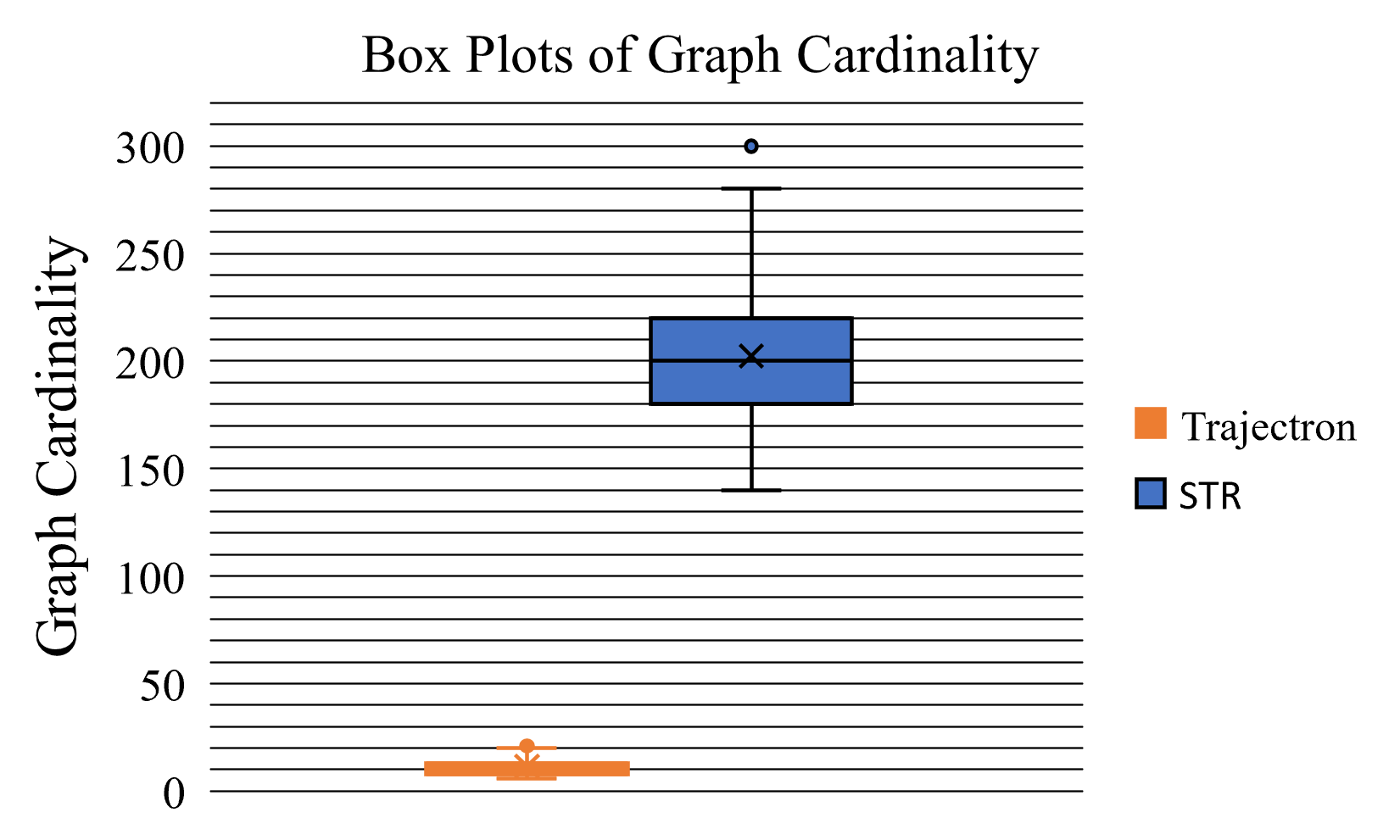}
    }
    
    \subfloat[75th percentiles density figure. x-axis displays percentage of non-zero edges out of 400 edges]{ \label{fig:7fth_perc}
    \includegraphics[width=\linewidth, height=5cm]{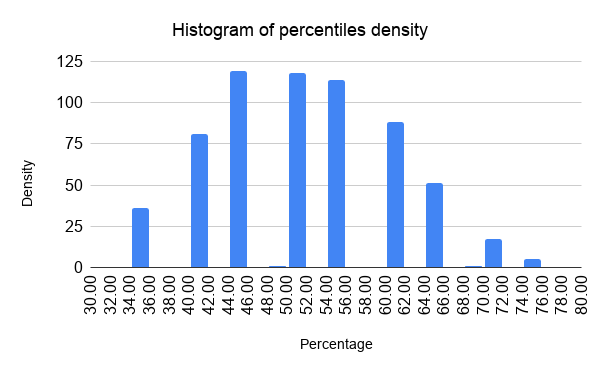}}
    
    \subfloat[The ratio of edge cardinality under STR to cardinality under fully-connected Ground-truth graph.]{\label{fig:str_to_fullgt}
    \includegraphics[width=\linewidth, height=5cm]{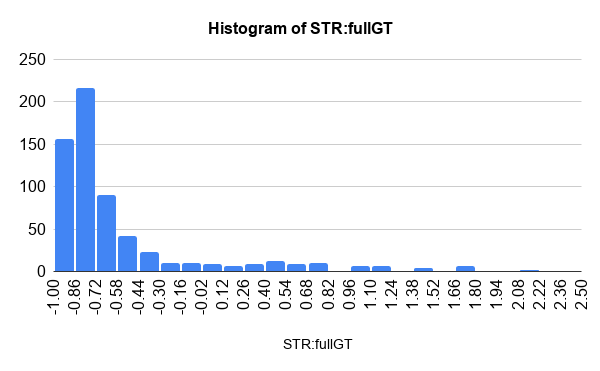}}
    \caption{Graph Cardinality. Statistical visualization of Self-Growing STR mechanism. }
    \label{fig:percentiles}
\end{figure}

Observing the whiskers boxplot in Figure \ref{fig:whisker_plt_1}, STR produces much lesser dense graphs than $SGTV$ \cite{haddad2020self} which yields a fully-connected (complete) graph. The y-axis represents edge counts (graph cardinality degrees). For a graph consisting 50 nodes at once, STR generates variational edge sets that range between 180 and 260 non-zero edges. Even though $SGTV$ generates weighted adjacency matrix, it only generates non-zero edges and therefore induces  2500 weighted edges. 

Figure \ref{fig:7fth_perc} shows the density of the 75th percentiles of STR to full$_{GT}$ graph cardinality. The highest density was found under 44\% and 52\% percentiles, which means that the
STR graph cardinality is smaller in size by almost half of the fully-connected graph case for a maximum of 20 pedestrians.
Figure \ref{fig:str_to_fullgt} shows that NMF construction of non-zero weighted edges was more efficient for higher pedestrian contexts. While the visualization is based on crowd size of 20 pedestrians, assuming a fully-connected graph is an inefficient design approach as in a big crowd pedestrian influence gets hindered by several factors such as crowdedness impact and limited visual awareness to immediate approaching pedestrians. Hence a fully-connected ground-truth graph is a partially applicable option as it can work for smaller crowd size. Over a crowd size of 9 pedestrians, STR assumes connectivity over 20 pedestrians. It ends up training a model with larger graph size than the ground-truth crowd. However, NMF generates sparse matrix that produces lesser falsely-assigned edges. 

For 9 pedestrians, STR to fullGT cardinality ratio was $\sim$ 2.50x. This means that STR generated 2.5x more edges.
For 10 pedestrians, STR generated 1x less edges, while for exactly 20 pedestrians, STR to fullGT cardinality ratio was down to 0.65x. 
For 50 pedestrians, STR ended up generating sparser graphs as it assumed that connectivity covers up to 20 pedestrians at one time-instance. The resultant graph ended up with $\sim$ 0.92x fewer edges compared to fullGT graph.
In percentage terms, at the first place, the reduction in graph cardinality was up to 86\% fewer edges over 200 STR proposal. At the second place, the reduction was up to 100\% fewer edges and this reduction was scored over nearly 150 STR proposals.

Figure \ref{fig:whisker_plt_1} measures the cardinality degree in Trajectron graph \cite{ivanovic2019trajectron}. It yields the least cardinality, given that it pools pedestrians based on the spatial proximity. This shows that Trajectron is more conservative in assigning edges between pedestrians, generating a minimal graph to model the social interactions. For a scene containing 15 pedestrians, Trajectron generates edge sets of size ranging between 6 and 20 edges, while STR generates a size ranging between 200 and 260 edges.

\section{STR Stability and Integrity}
The proposed framework deploys the Nonnegative Matrix Factorization to be the edges construction mechanism. It works as a recommender system for completing the online Adjacency matrix, which is realized by taking the downscaled approximated features $W$ (assigned an initial attention matrix), and the components part $H$ (assigned the hidden states of pedestrians). 
Since this neural reconstruction is a divergent process and that the optimal adjacency matrix $A^*$ is not observable, the Nonnegative Matrix Factorization of the best adjacency is an NP-Hard problem that yields local optimal adjacency modeling. Similar case demonstrating the graph-structured NP-hard problems appears in \cite{joshi2019efficient} which exploits graph neural traversal given a complete set of edges that is weighted as a local optimum adjacency matrix. In our graph case, the output will be the reconstructed attention weights $W'$ and hidden states $H'$. $W'$ is scaled to 1 and used as the weighted adjacency matrix $A_0$ to represent the weighted social interaction.

In the following lemma, the adjacency matrix (A) is the latent variable since it is not fully-observable. The priors, features matrix W (represents the edges attention weight) is assigned ($\hat{F}$), and the components matrix H is assigned ($C_{map}$) are fully-observable and set a-priori, so the reconstruction of $A$ is conditioned by $H$, $W$ and a nonnegative $A_0$ that is initialized to ones. The adjacency at any index and at any instance in time remains conditioned by the priors.

The Null hypothesis $h_0$ indicates that bounding matrix $A$ by a distribution-free distribution $H$ leads at the end to a Normal distribution:
\begin{equation}
h_0: \quad A_{(i,j)} \in P(W_{(i,j)}|H_{(i.j)},A_0) \sim G(\mu=0, \sigma=1)
\end{equation}

The Null hypothesis $h_0$ also indicates that matrix $A$ always obey some parametric distribution which can be conveyed as a Gaussian distribution $G(\mu=0, \sigma=1)$.

On the other hand, the alternative hypothesis $h_1$ says that NMF generates variational results which cannot be conveyed by the normality assumption. Therefore, the distribution-free assumption represents a wider set of distribution that can better model the social interaction. 

$h_1$ represents a more generic modeling of pedestrians adjacency (occupancy) states, of which the Gaussian distribution bounding conveys a limited part of the social interaction modeling:
\begin{equation}
h_1: \quad A_{(i,j)} \in P(W_{(i,j)}|H_{(i.j)},A_0) \sim H
\end{equation}

\begin{figure}
	\includegraphics[width=\linewidth, height=6cm]{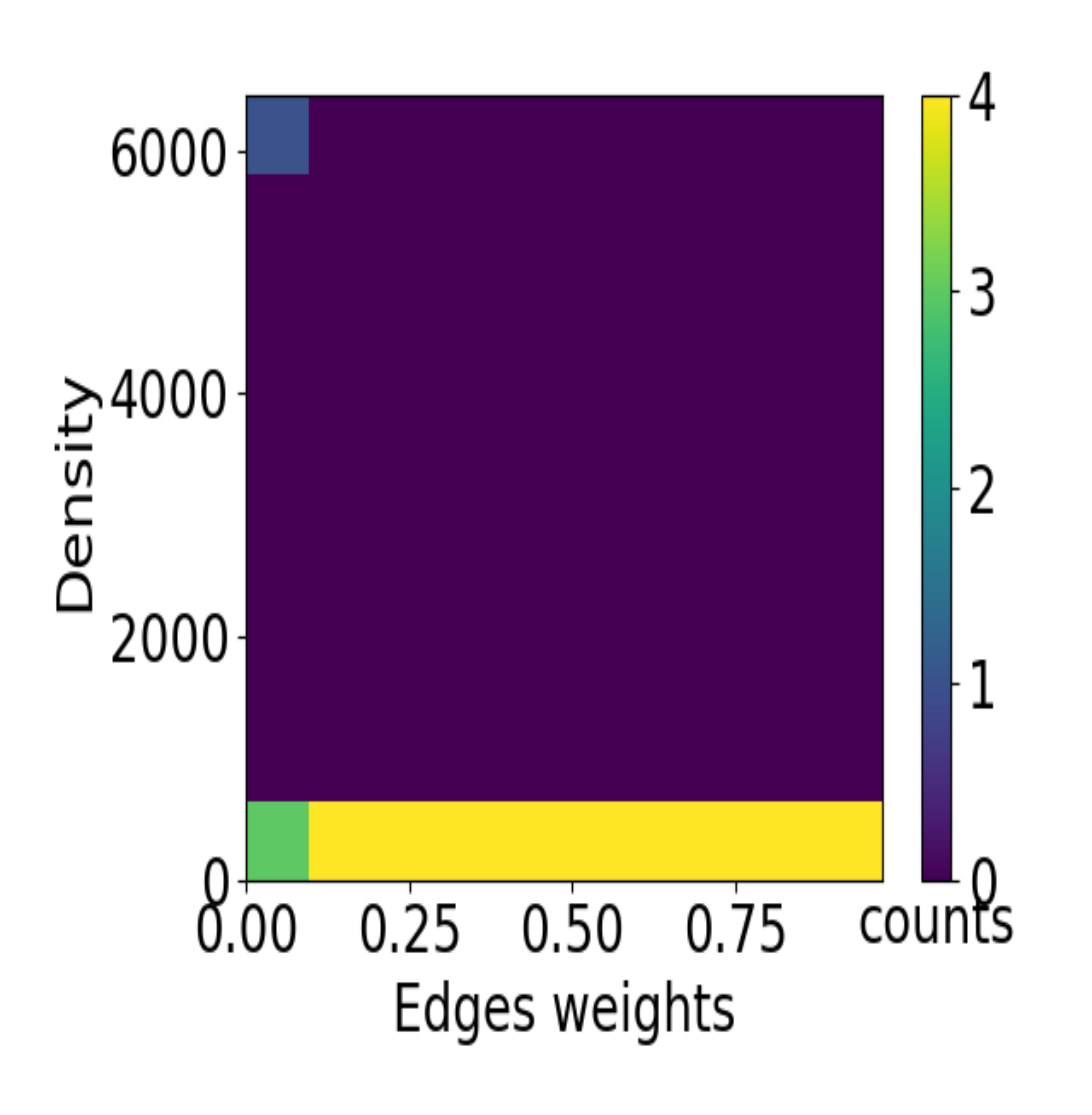}
	\caption{Density histogram of Adjacency weights, designates graph edges density and weight concentration values.}
	\label{app:hmaps}
\end{figure}

By visually observing the Density function in Figure \ref{app:hmaps}, a plot of Adjacency matrices over 400 runs of NMF algorithm, the density values indicate mostly sparse matrices with the highest density of 0s and a narrow span of nonzero weights that have $\le 2000$ occurrences, which gives the impression of normally-based adjacency and supports the Null hypothesis $h_0$. 

This designates that all estimated adjacencies by NMF will always obey the normality condition, such that the estimated values will always be small and the density function is concentrated around zero. The probability of any reconstructed features using NMF obey a univariate Gaussian distribution.

\begin{figure*}
\centering
	\subfloat[Average Displacement curves under GGRNN model over the 1st epoch round of online training. Errors are reported in \textbf{meters}.]{ \label{oc_eth}
	\includegraphics[width=9cm, height=5cm]{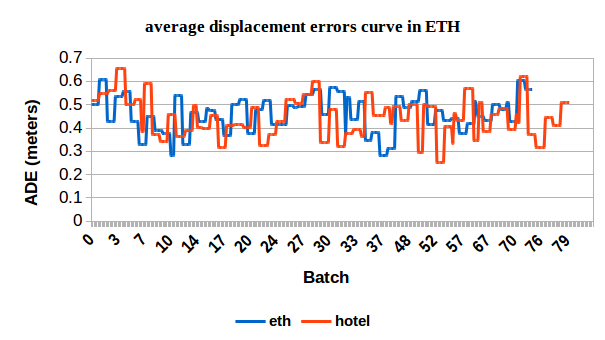}}
	\subfloat[Average Displacement curves under GGRNN-V model over the 1st epoch round of       online training. Errors are reported in \textbf{meters}.]{ \label{ovc_ucy}
	\includegraphics[width=9cm, height=5cm]{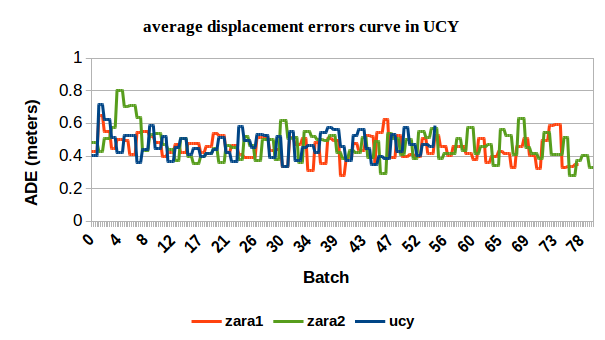}}
	\caption{ADE curves under GGRNN and GGRNN-V models shows the step-wise function pattern over ETH-UCY dataset. The step-wise pattern implies that Gaussian-based assumption may not be the best and generic distribution to fit the predictive model patterns. Therefore, distribution-free hypothesis is a more proper assumption for trajectory predictions generated by our models.}
\label{fig:ovc_crowds}
\end{figure*}

\begin{figure}
\centering
    \subfloat[STR-GGRNN curves]{ \label{oc_deathcircle}
    \includegraphics[width=\linewidth, height=5cm]{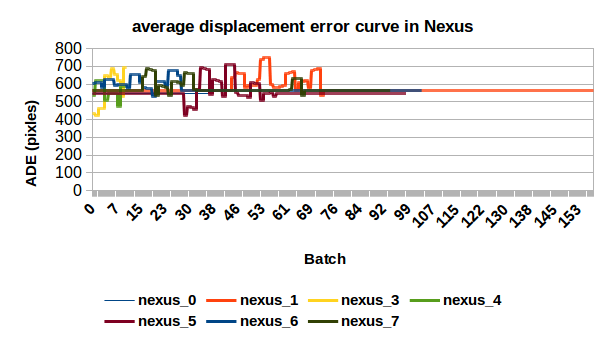}}\\
    \subfloat[GGRNN curves]{ \label{oc_nexus}
    \includegraphics[width=\linewidth, height=5cm]{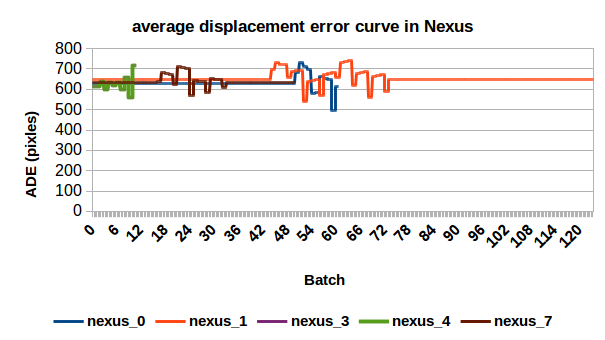}}\\
    \subfloat[STR-GGRNN curves]{ \label{osc_hyang}
    \includegraphics[width=\linewidth, height=5cm]{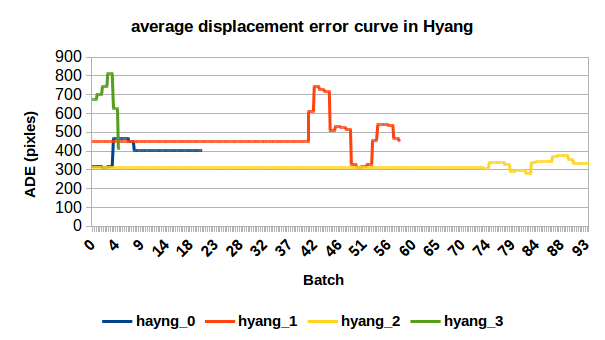}}\\
    \subfloat[GGRNN curves]{ \label{oc_hyang}
    \includegraphics[width=\linewidth, height=5cm]{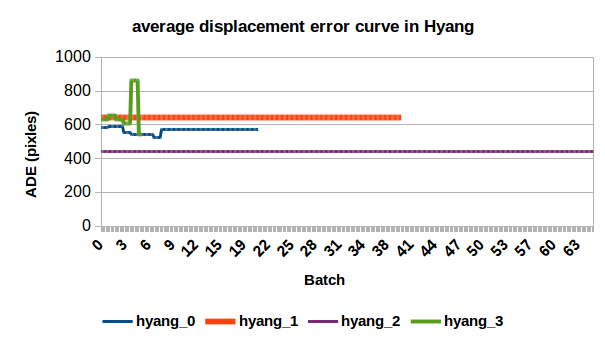}}
    \caption{ADE curves reflect the step-wise and stable patterns over SDD. These suggest that the proposed predictor model behavior cannot be fit under Gaussian distribution and is best described as a model-free method. Distribution-Free assumption for such a predictor system is a more generic hypothesis to describe the model performance.}
    \label{fig:oc_ade}
\end{figure}

\begin{figure}
\centering
    \subfloat[STR-GGRNN curves]{ \label{osc_deathcircle0}
    \includegraphics[width=\linewidth, height=5cm]{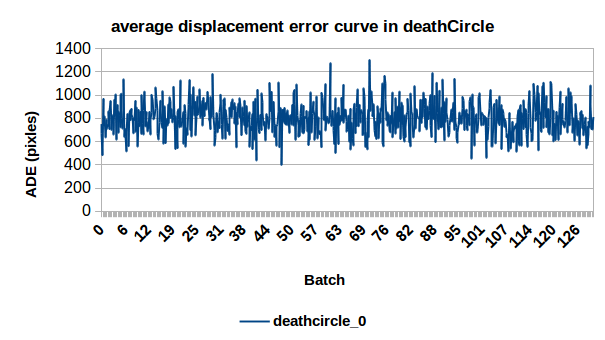}}\\
        \subfloat[GGRNN curves]{ \label{oc_deathcircle_1}
    \includegraphics[width=\linewidth, height=5cm]{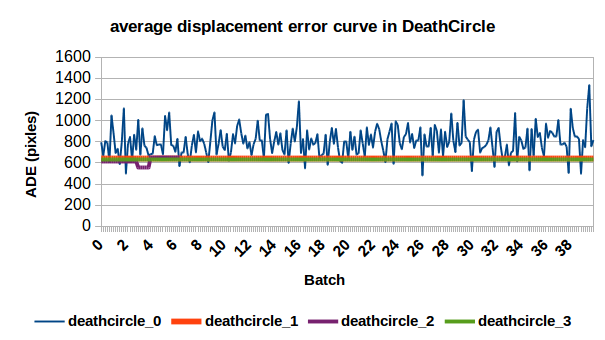}}
    \caption{Average Displacement curves under ST-GGRNN model over the 1st epoch round of online training. Euclidean errors are reported in Pixels. The ADE curves are generated at frames of full resolution for Death Circle 4 sets.}
    \label{fig:osc_ade_curves_sdd_deathcircle}
\end{figure}

On the other hand, observing the loss curves reveals what can be a more convenient distribution of pedestrian trajectory data at its entirety.
As the loss function is set as L2-Norm, the ADE curves in Figures \ref{fig:ovc_crowds},\ref{fig:osc_ade_curves_sdd_deathcircle},\ref{fig:oc_ade} can be considered for analyzing the predictor models performance on one side, and on the other side, they can be used to analyze the regression pattern. 

The STR-GGRNN ADE curves regressions are normally distributed, while the GGRNN-V curves regression pattern designates a \textit{levy-stable} distribution. The latter distribution is clearly designated by GGRNN and ST-GGRNN model under the SDD training curves.

This variation in predictive model regressions, shows that the normality assumption cannot apply to all the tested contexts. Under SDD, the pattern shows a range of step-like, stable, and normal regressions, while under ETH-UCY, there were consistent appearance of normal regressive curves.

Overall, the proposed model was trained as a model-free method by taking updates immediately. There are no assumptions made before training and estimation about the distribution parameters/type that convey the adjacency modeling, nor there was any parametric mapping of hidden states onto a parameters vector \cite{alahi2016social, vemula2018social}, therefore, the proposed predictive models performance is better understood under distribution-free hypothesis. In this case, the alternative hypothesis $h_1$ can provide a more generic probabilistic perception of the predictive model. 

While STR-GGRNN-V performance obeys the normal distribution pattern, other model variants represented different patterns that are outside the normal distribution category. As indicated by the predictor model performance in Figures \ref{oc_eth} -- \ref{oc_hyang}. These figures show stable function and step-wise function curves in STR-GGRNN and GGRNN models.
In the literature, the normality assumption was always set as the prior knowledge to eliminate the prediction problem and ensure the predictions integrity. That is to simplify the probabilistic modeling under dynamic interactive systems.
Now, it is safe to conclude that the normality assumption (Null hypothesis $h_0$) is partially applicable to the predictor systems application. It is proper to assume distribution-free prediction patterns on pedestrian trajectory to perceive the predictors performance. Considering that STR-GGRNN and STR-GGRNN-V models includes an unsupervised adjacency estimation using NMF and it follows the Null hypothesis, the STR-GGRNN model yields a prediction behavior that is non-Gaussian. This means the model-free techniques can be modeled using a probability distribution that is not necessarily Gaussian over the human trajectory all the time. So to have a general perspective, the trajectory prediction model can be formulated without a specific distribution assumption assumption and generate useful predictions.


\section{Conclusion}
In this work, we presented STR-GGRNN, a framework for adaptive social relational inference, through online nonnegative matrix factorization as a data-driven graph-completion task. Our model mixes local contextual modeling with the global social modeling of crowd interactions without relying on a predefined proxemics threshold. NMF is an NP-hard problem, so theoretically, it is proven to reach a sub-optimal solution of the best neighboring recommendation. Currently, the approach selects best adjacency with log-polynomial time-complexity. Integrating social inference given rich enhanced interactions and neighborhoods representation, constitutes for prediction accuracy outcomes and withdraws the exhaustive search of optimal neighborhood problem.
Experimental results showed significant improvement in future trajectories prediction. It reduced errors down to 10cm on average. 
Although the adjacency matrix assumes a fixed scene size, the model is proxemic-free and the adjacency size is fitting for a moderately-crowded environment.
As future work, we plan to model adjacency directly on the graph without mapping to fixed-size matrices. We also plan to extend the testing to heterogeneous crowd graph given the learned knowledge from homogeneous crowds, wherein the adaptive online learning has higher potential.

\bibliographystyle{IEEEtran.bst}
\bibliography{IEEEexample.bib}

\end{document}